\documentclass{article}

\usepackage[final]{neurips_data_2024}

\usepackage{pifont}
\usepackage[dvipsnames,table]{xcolor}         
\usepackage{wrapfig}
\usepackage[
    pagebackref=true,
    colorlinks=true,
    urlcolor=MidnightBlue,
    linkcolor=MidnightBlue,
    citecolor=MidnightBlue]
{hyperref}       
\usepackage{url}            
\usepackage{booktabs}       
\usepackage{amsmath}
\usepackage{amsfonts}       
\usepackage{nicefrac}       
\usepackage{microtype}      
\usepackage{graphicx}
\usepackage{bm}
\usepackage{listings}
\usepackage{longtable}

\usepackage{pythonhighlight}
\usepackage{cleveref}
\crefname{lstlisting}{listing}{listings}
\Crefname{lstlisting}{Listing}{Listings}
\usepackage[utf8]{inputenc} 
\usepackage[T1]{fontenc}    
\usepackage{float}

\newfloat{lstfloat}{htbp}{lop}
\floatname{lstfloat}{Listing}

\definecolor{codegreen}{rgb}{0,0.6,0}
\definecolor{codegray}{rgb}{0.5,0.5,0.5}
\definecolor{codepurple}{rgb}{0.58,0,0.82}
\definecolor{backcolour}{rgb}{0.95,0.95,0.92}
\definecolor{summarygray}{rgb}{0.5, 0.5, 0.5}

\lstdefinestyle{mystyle}{
    keywords={if, for, procedure, or, else, return},
    backgroundcolor=\color{backcolour},   
    commentstyle=\color{codegreen},
    keywordstyle=\color{magenta},
    numberstyle=\tiny\color{codegray},
    stringstyle=\color{codepurple},
    basicstyle=\ttfamily\scriptsize,
    breakatwhitespace=false,         
    breaklines=true,                 
    captionpos=b,                    
    keepspaces=true,                 
    showspaces=false,                
    showtabs=false,                  
    tabsize=2
}
\renewcommand{\paragraph}[1]{\textbf{#1}~~}  

\newcommand{\poli}{\texttt{poli}}
\newcommand{\polibaselines}{\texttt{poli-baselines}}

\title{A survey and benchmark of high-dimensional \\ Bayesian optimization of discrete sequences}

\author{%
  Miguel González-Duque\thanks{emails: \texttt{miguelgondu@gmail.com}, \texttt{sohau@dtu.dk}, \texttt{wb@di.ku.dk}}\\
  University of Copenhagen\\
  \And
  Richard Michael\\
  University of Copenhagen\\
  \AND
  Simon Bartels\\
  Paul Sabatier University Toulouse\\
  \And
  Yevgen Zainchkovskyy\\
  Technical University of Denmark\\
  \And
  Søren Hauberg$^*$\\
  Technical University of Denmark\\
  \And
  Wouter Boomsma$^*$\\
  University of Copenhagen\\
}

\begin{document}

\maketitle
    
\begin{abstract}
Optimizing discrete black box functions is key in several domains, e.g.\@ protein engineering and drug design. Due to the lack of gradient information and the need for sample efficiency, Bayesian optimization is an ideal candidate for these tasks. 
Several methods for high-dimensional continuous and categorical Bayesian optimization have been proposed recently.
However, our survey of the field reveals highly heterogeneous experimental set-ups across methods and technical barriers for the replicability and application of published algorithms to real-world tasks. To address these issues, we develop a unified framework to test a vast array of high-dimensional Bayesian optimization methods and a collection of standardized black box functions representing real-world application domains in chemistry and biology. These two components of the benchmark are each supported by flexible, scalable, and easily extendable software libraries (\texttt{poli} and \texttt{poli-baselines}), allowing practitioners to readily incorporate new optimization objectives or discrete optimizers. Project website: \url{https://machinelearninglifescience.github.io/hdbo_benchmark}.
\end{abstract}

\section{Introduction}

Optimizing an unknown and expensive-to-evaluate function is a frequent problem across disciplines \citep{Shahriari2016BOreview}, examples are finding the right parameters for machine learning models or simulators, drug discovery \citep{Bombarelli:AutoChemDesign:2018,Griffiths:ConstrainedBOVAEs:2020,Pyzer-Knapp:BODrugDiscovery:2018}, protein design \citep{Stanton:LAMBO:2022,Gruver:LAMBO2:2023}, hyperparameter tuning in Machine Learning \citep{Snoek:PracticalBO:2012,Turner:BOHyperTuning:2020} and train scheduling. In some scenarios, evaluating the black box involves an expensive process (e.g. training a large model, or running a physical simulation); Bayesian Optimization (BO, \citet{Mockus:OriginalBO:1975}) is a powerful method for sample efficient black box optimization.
High dimensional (discrete) problems have long been identified as a key challenge for Bayesian optimization algorithms \citep{Wang2013rembo,Snoek:PracticalBO:2012} given that they tend to scale poorly with both dataset size and dimensionality of the input. 

\begin{figure}
    \centering
    \includegraphics[width=.95\textwidth]{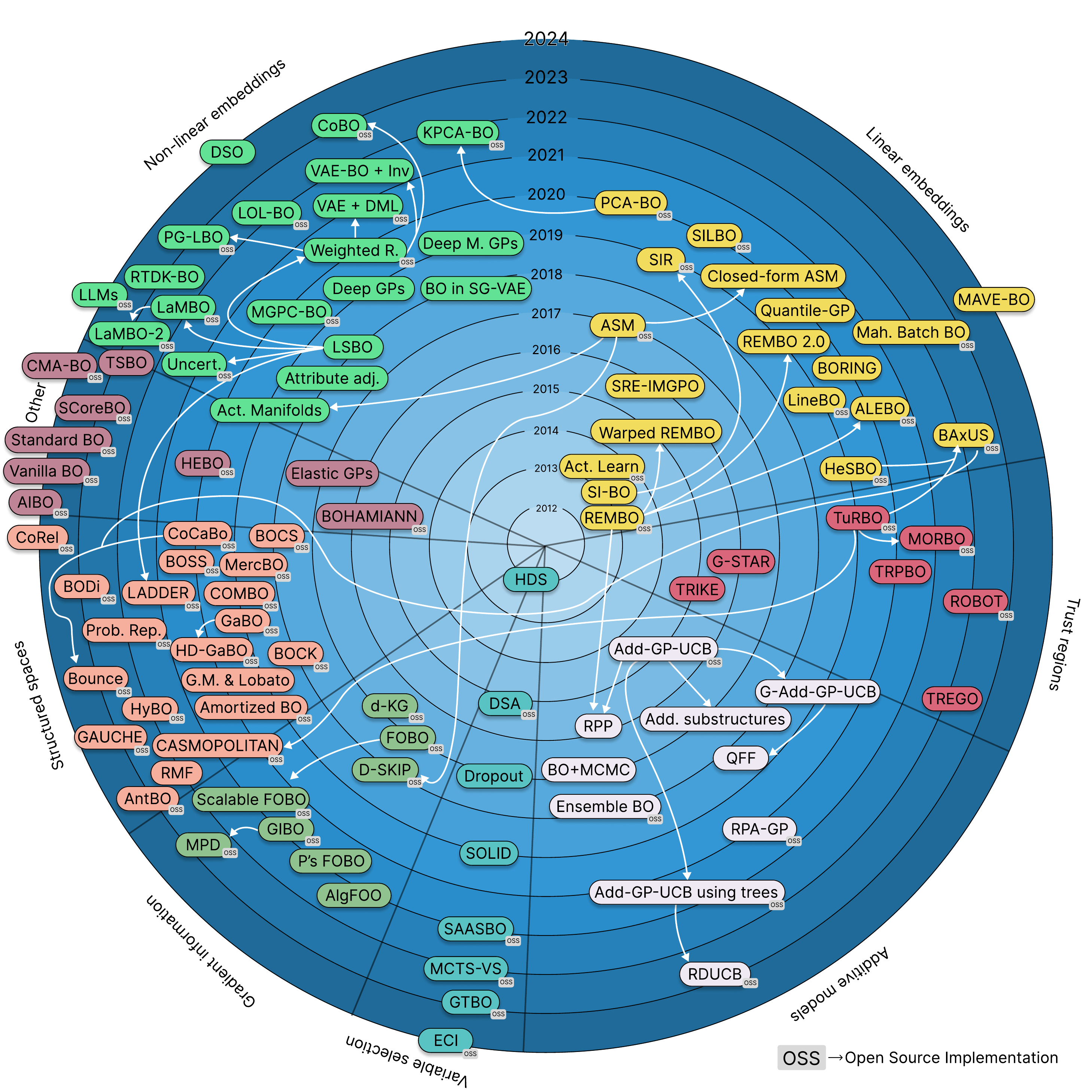}
    \caption{A timeline of high-dimensional Bayesian optimization methods, with arrows drawn between methods that explicitly augment or use each other. References can be found in supplementary Table~\ref{tab:appendix:full_taxonomy}. The figure is inspired by \citet{Justesen:DL4VGP:2020}. \href{https://machinelearninglifescience.github.io/hdbo_benchmark/docs/hdbo/introduction/}{An interactive version can be found in our project page}.}
    \label{fig:hdbo_timeline}
\end{figure}

\begin{figure}
    \centering
    \includegraphics[width=0.95\textwidth]{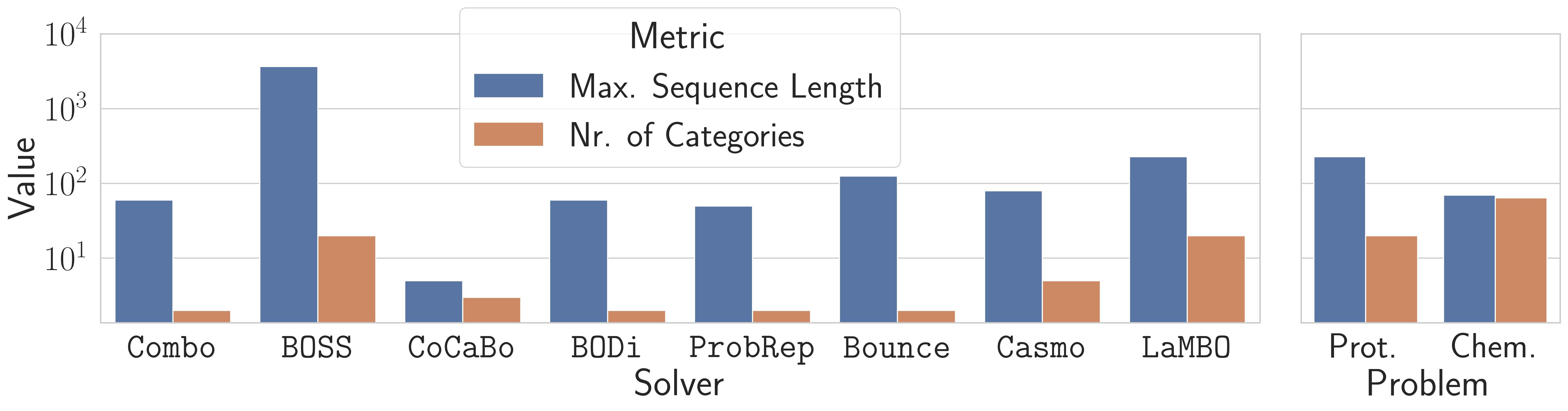}
    \caption{Existing BO methods tackle problems with insufficiently low effective dimensions. This figure shows sequence length and nr.\@ of categories of the highest search space in the original tests. For reference, the discrete optimization problems usually tackled by practitioners in chemistry and biology are of the order of $10^2$ in sequence length, and $>10^1$ in nr.\@ of categories. Methods that optimize directly in discrete space (e.g.\@ \texttt{BODi}, \texttt{ProbRep}, \texttt{Bounce}; Sec.~\ref{sec:taxonomy:structured-spaces}) are tested in lower sequence lengths and dictionary sizes; methods that rely on unsupervised information (e.g.\@ \texttt{LaMBO}, etc.;  Sec.~\ref{sec:taxonomy:non-linear-embeddings}) are able to optimize more complex problems, like protein engineering or small molecule optimization.}
    \label{fig:overview-of-effective-dim}
\end{figure}

High-dimensional BO has been the focus of an entire research field (see Fig.~\ref{fig:hdbo_timeline}), in which methods are extended to address the curse of dimensionality and its consequences \citep{BinoisWycoff:HDGPs:2022,SantoniDoerr:HDBO:2023}.
Within this setting, discrete sequence optimization has received particular focus, due to its applicability in the optimization of molecules and proteins. However, prior work often focuses on sequence lengths and number of categories below the hundreds (see Fig. \ref{fig:overview-of-effective-dim}), making it difficult for practitioners to judge expected performance on real-world problems in these domains.
We contribute (i) a survey of the field while focusing on the real-world applications of high-dimensional discrete sequences, (ii) a benchmark of several optimizers on established black boxes, and (iii) an open source, unified interface: \texttt{poli} and \texttt{poli-baselines}.\footnote{\texttt{https://github.com/MachineLearningLifeScience/\{\href{https://github.com/MachineLearningLifeScience/poli}{poli}, \href{https://github.com/MachineLearningLifeScience/poli-baselines}{poli-baselines}\}}}

\section{Preliminaries}


\paragraph{Bayesian Optimization and Gaussian processes.}
Bayesian optimization requires a surrogate model and an acquisition function \citep{Garnett2023BObook}.
Given both, the objective function is sequentially optimized by fitting a model to the given observations and numerically optimizing the acquisition function with respect to the model to select the next configuration for evaluation.  
Frequently, the model is a Gaussian process (GP, \citet{RasmussenWilliams:GPs:2006}), and popular choices for the acquisition function are \emph{Expected Improvement} \citep{Jones:BO:1998,Garnett2023BObook} and the \emph{Upper Confidence Bound} \citep{Srinivas:UCB:2012}. 
A GP allows to express a prior belief over functions. 
Formally, it is a collection of random variables, such that every finite subset follows a multivariate normal distribution, described by a mean function $\mu$, and a positive definite covariance function (kernel) \citep[p.~13]{RasmussenWilliams:GPs:2006}. 
Assuming that observations of the function are distorted by Gaussian noise, the posterior over the function conditioned on these observations is again Gaussian.
The prediction equations have a closed form and can be evaluated in $\mathcal{O}(N^3)$ time where $N$ is the number of observations.





\paragraph{Is high-dimensional Bayesian Optimization difficult?}
There are three reasons why BO is thought to scale poorly with dimension. The first reason is that GPs fail to properly fit to the underlying objective function in high dimensions. Secondly, even if the GPs were to fit well there is still the problem of optimizing the high-dimensional acquisition function. Finally, Gaussian Processes are believed to scale poorly with the size of the dataset, limiting us to low-budget scenarios \citep{BinoisWycoff:HDGPs:2022}. Folk knowledge suggests that GPs fail to fit functions above the meager limit of $\sim 10^1$ dimensions \citep{SantoniDoerr:HDBO:2023} and $\sim 10^4$ datapoints.

\citet{Hvarfner:VanilaBO:2024} recently disputed these well-entrenched narratives by showing that poor GP fitting could be caused by a poor choice of regularizer; mitigating the curse of dimensionality could be as easy as including a dimensionality-dependent prior over lengthscales. Furthermore, \citet{Xu:StdVanillaBO:2024} argues that even the simplest BO outperforms highly elaborate methods.


\paragraph{Optimization of discrete sequences \& applications.}
Most HDBO methods are tested on toy examples, hyperparameter tuning, or reinforcement learning tasks 
\citep{BinoisWycoff:HDGPs:2022,Penubothula:FirstOrder:2021}. 
We focus on discrete sequence optimization, which has
several applications beyond the usual examples (e.g.\@ MaxSAT, or PestControl)
\citep{Papenmeier:BOUNCE:2024},
and is key in applications to biology and bioinformatics \citep{Bombarelli:AutoChemDesign:2018, Stanton:LAMBO:2022, Gruver:LAMBO2:2023}. 
Drug design and protein engineering can be thought of as sequence optimization problems, if we consider the SMILES/SELFIES representation of small molecules \citep{Weininger:SMILES:1988, Krenn:SELFIES:2020}, or the amino acid sequence representation of proteins \citep{Needleman:AminoAcids:1970}.

\paragraph{Related work.} 
\citet{BinoisWycoff:HDGPs:2022} initially surveyed the field of high-dimensional GPs, focusing on applications to BO, and proposed a taxonomy of structural assumptions for GPs that includes \textit{variable selection}, \textit{additive models}, \textit{linear}, and \textit{non-linear embeddings}. 
This work has since been updated by \citet{Wang:BOAdvances:2023} and \citet{SantoniDoerr:HDBO:2023}. The latter presents an empirical study on the continuous, toy-problem setting up to 60 dimensions and refines the taxonomy \citep{BinoisWycoff:HDGPs:2022} into five categories, separating \textit{trust regions} from the rest.
\citet{Griffiths:GAUCHE:2023} compare different kernel functions used with binary vector representations of molecules and for the same application \citet{Kristiadi:LLM:2024} study the use of different large language models.
Our work is most similar to \citet{dreczkowski:mcbo:2023}'s comprehensive overview of discrete BO (\texttt{MCBO}), and \citet{Gao:PMOMolOpt:2022}'s benchmark of small molecule optimization. In both, HDBO is not in focus.

\section{A taxonomy of high-dimensional Bayesian Optimization}
\label{sec:taxonomy}


We describe the field of high dimensional BO and the large number of related publications through a refined taxonomy building on previous work, discussing \emph{variable selection}, \emph{additive models}, \emph{trust regions}, \emph{linear embeddings}, \emph{non-linear embeddings}, \emph{gradient information}, \emph{structured spaces}, and others in turn. 
While encompassing taxonomies over fields may initially appear ill-advised \citep[pp.22]{Wilkins:PhilLang:1668},
we highlight commonalities in strategies that give structure to the HDBO problem-space.

We expand previous surveys \citep{BinoisWycoff:HDGPs:2022, SantoniDoerr:HDBO:2023} and identify a finer taxonomy of seven method groups and new families of \textit{structured spaces} (i.e.\@ methods that work directly on mixed representations, or Riemannian manifolds, previously categorized as \textit{non-linear embeddings}), and methods that rely on predicted \textit{gradient information}. This new separation emphasizes the heterogeneous nature of discrete solvers: some optimizers work directly on discrete space (\textit{structured spaces}), while others optimize using latent representations (\textit{non-linear embeddings}); gradient-based methods are separated to show alternatives when first-order information is available or modelable.
Fig.~\ref{fig:hdbo_timeline} presents a timeline of HDBO methods, split into these families, and all methods are detailed in supplementary Table~\ref{tab:appendix:full_taxonomy}; methods are grouped according to their most dominant feature.


\subsection{Variable selection}

To solve a high-dimensional problem, one approach is to focus on a subset of variables of high interest.\footnote{Under the assumption that there exists an axis-aligned lower-dimensional \textit{active subspace}.} 
One selects the variables either by using domain expertise, or by Automatic Relevance Detection (ARD) \citep[pp.106-107]{RasmussenWilliams:GPs:2006} i.e.\@ large lengthscales indicate independence under the covariance matrix for GPs.
%
Examples of this approach include Hierarchical Diagonal Sampling (HDS) \citep{Chen:HDS:2012} and the Dimension Scheduling Algorithm (DSA) \citep{Ulmasov:DSA:2016}. The former determines the active variables by a binary tree of subsets of $\{1, \dots, D\}$, and fits GPs in lower-dimensional projections. 
DSA constructs a probability distribution by the principal directions of the training inputs $\{(\bm{x}_n, y_n)\}_{n=1}^N$ and subsamples the dimensions accordingly.
In contrast \citet{Li:HDBODropout:2018} randomly sample subsets of active dimensions.

Other methods rely on placing priors on their lengthscales, followed by a Bayesian treatment of the training. 
In Sequential Optimization of Locally Important Directions (SOLID), lengthscales are weighted by a Bernoulli distributed parameter, and coordinates are removed when their posterior probability goes below a user-specified threshold.
\citep{Munir:SOLID:2021}. \citet{Eriksson:SAASBO:2021} consider the Sparse Axis-Aligned Subspace (SAAS) model of a GP, restricting the function space through a (long-tailed) half-Cauchy prior on the inverse-lengthscales of the kernel. 

\subsection{Additive models}

Additive models assume that the objective function $f$ can be decomposed into a sum of lower-dimensional functions. 
Symbolically, the coordinates of a given input $\bm{x} = (x_1, \dots, x_D)$ are split into $M$ usually disjoint subgroups $g_1, \dots g_M$ of smaller size, called a decomposition. 
Instead of fitting a GP to $D$ variables in $f$, the algorithm fits $M$ GPs to the restrictions $f|_{g_1}, \dots f|_{g_M}$ and adds their Upper Confidence Bound. 
The differences between the algorithms in this family are on how the subgroups are constructed, how the additive structure is approximated, the training of the Gaussian Process, or leveraging special features \citep{Mutny:HDBOQFF:2018}.

\citet{Han:AddGPUCB:2021} select the decomposition which maximizes the marginal likelihood from a collection of randomly sampled decompositions, updating it every certain number of iterations. 
Alternatives include: leveraging a generalization based on restricted projections \citep{Li:RPP:2016}, discovering the additive structure using model selection and Markov Chain Monte Carlo \citep{Gardner:AddStruct:2017}, considering overlapping groups \citep{Rolland:GAddGPUCB:2018}, ensembles of Mondrian space-tiling trees \citep{Wang:BatchedEnsembleHDBO:2018}, 
or use random tree-based decompositions \citep{Ziomek:RDUCB:2023}. 

\subsection{Trust regions}

Some BO algorithms restrict the evaluation of the acquisition function to a small region of input space called a \textit{trust region}, which is centered at the incumbent and is dynamically contracted or expanded according to performance \citep{Rommel:TRIKE:2016,Pedrielli:GSTAR:2016, Eriksson:TuRBO:2019}.
Contemporary variants extend to the multivariate setting (e.g.\@ MORBO \citep{Daulton:MORBO:2022}), to quality-diversity \citep{Maus:ROBOT:2023} and to the optimization of mixed variables (\texttt{CASMOPOLITAN} by \citet{Wan:CASMOPOLITAN:2021}), including categorical.
Since the trust region framework involves only the optimization of the acquisition function, several other methods leverage it alongside other structural assumptions like linear/non-linear embeddings (e.g.\@ \citet{Tripp:WeightedRetraining:2020, Papenmeier:BAxUS:2022}).

\subsection{Linear embeddings}
\label{sec:taxonomy:linear_embeddings}


Instead of optimizing directly in input space $\mathbb{R}^D$, several methods rely on optimizing in a lower-dimensional space $\mathbb{R}^d$, which is linearly embedded into data space using a linear transformation $A\in\mathbb{R}^{D\times d}$ \citep{Wang:REMBO:2016}. The matrix $A$ can be either selected at random \citep{Wang:REMBO:2016, Qian:SREIMGPO:2016}, computed as a low-rank approximation of the input data matrix \citep{Djolonga:HDBandits:2013, Zhang:SIR:2019,Raponi:PCABO:2020}, constructed using gradient information and active subspaces \citep{Palar:ASM:2017, Wycoff:AS:2021}, or through the minimization of variance estimates \citep{Hu:MAVEBO:2024}.

These methods are limited by how low-dimensional exploration translates into high dimensions. 
One choice of embedding matrix $A$ spans a \textit{fixed}, highly-restricted subspace of $\mathbb{R}^D$. 
For this approach several issues regarding back-projections need to be addressed. Indeed, projecting from bounded domains $\mathcal{Z}\subseteq\mathbb{R}^d$ to $\mathbb{R}^D$ might render points outside the bounded domain in the input \citep{BinoisWycoff:HDGPs:2022}. 
Finally, the transformation $A$ is not injective, meaning a point in input space can correspond to several latent points \citep{Binois:WarpedREMBO:2015, Moriconi:QuantGPBO:2020}.

\citet{Binois:WarpedREMBO:2015} propose a kernel that alleviates these issues by including a back-projection to the bounded domain that respects distances in the embedded space.   
 \textit{Hashing matrices} $S\in\mathbb{R}^{D\times d}$ are an alternative way to reconstruct an input data point in a bounded domain $\bm{x}\in [-1, 1]^{D}\subseteq\mathbb{R}^D$ from a latent point $\bm{z}\in\mathbb{R}^D$, whose entries are either 0, 1, and -1. 
Thus, the result of multiplying $S\bm{z}$ is a linear combination of the coordinates of $\bm{z}$ where the coefficients are 1 and -1 \citep{Nayebi:HESBO:2019}. 
These ideas have been combined with trust regions both in the continuous \citep{Papenmeier:BAxUS:2022} and mixed-variable settings \citep{Papenmeier:BOUNCE:2024}. 
A natural extension considers a family of nested subspaces, progressively growing the embedding matrix until it matches the input dimensionality \citep{Papenmeier:BAxUS:2022}. 
An alternative that does not deal with reconstruction mappings (thus circumventing the aforementioned issues) uses the information learned in the lower dimensional space to perform optimization directly in input space \citep{Horiguchi:MahalaBatchBO:2022}.

\subsection{Non-linear embeddings}
\label{sec:taxonomy:non-linear-embeddings}



Several methods have considered non-linear embeddings to incorporate learned latent representations.
One set of examples are deep latent variable models like Generative Adversarial Networks \citep{Goodfellow:GAN:2014}, or variants of Autoencoders \citep{Kingma:VAE:2014,Stanton:LAMBO:2022,Maus:LOLBO:2022}, algorithms that allow for modelling arbitrarily structured inputs. This is highly relevant for optimizing sequences, which are modeled as samples from a categorical distribution.


\citet{Bombarelli:AutoChemDesign:2018} pioneered latent space optimization (LSBO) by learning a latent space of small molecules through their SMILES representation using a Variational Autoencoder (VAE, \citet{Kingma:VAE:2014,Rezende:VAE:2014}), and optimizing metrics such as the qualitative estimate of druglikeness (QED) therein. Several approaches have followed, including usage of \emph{a-priori} given labelled data \citep{Eissman:AttrAdjust:2018} or decoder uncertainty \citep{Notin:Uncert:2021},
smart retraining schemes that focus on promising points \citep{Tripp:WeightedRetraining:2020}, metric-learning approaches that match promising points together \citep{Grosnit:VAEDML:2021}, 
constraining the latent space \citep{Griffiths:ConstrainedBOVAEs:2020},
latent spaces mapping to graphs \citep{Kusner:GrammarVAE:2017,Jin:JunctionTreeVAE:2018}
and jointly learning the surrogate model and the latent representation \citep{Maus:LOLBO:2022,Lee:CoBo:2023, Chen:PGLBO:2024, Kong:DSBO:2024}. 
\citet{Stanton:LAMBO:2022} take this further by learning multiple representations: one shared and required for both the decoder and surrogate, and one discriminative encoding as input for a GP used in the acquisition function.
A prerequisite for these methods is a large dataset of \textit{unsupervised} inputs, which may not be available in all applications. 
The methods that rely on training both the representation and the regression at the same time need \textit{supervised} labels, which may be potentially unavailable. 
Optimization in embedding spaces greatly increases the complexity of problems that can be tackled, making it an appealing alternative for discrete sequence optimization in real-world tasks (see Fig.~\ref{fig:overview-of-effective-dim}).

\subsection{Gradient information}


High-dimensional problems can become significantly easier when derivative information is available.
Even when the objective's derivatives are not available, the gradient information from the surrogate model can guide exploration.
In our case, the referenced approaches cannot be applied directly, as they assume a differentiable kernel. For methods that rely on a continuous latent representation (see Secs.~\ref{sec:taxonomy:linear_embeddings} and \ref{sec:taxonomy:non-linear-embeddings}), gradient information of the surrogate model in latent space can be used.

\citet{Ahmed:FOBO:2016} mention how several Bayesian optimization methods could leverage gradient information and encourage the community to augment their optimization schemes with gradients, supported by strong empirical results even with randomly sampled directional derivatives. \citet{Eriksson:DSKIP:2018} alleviate the computational constraints that come from using supervised gradient information using structured kernel interpolation and computational tricks like fast matrix-vector multiplication and pivoted Cholesky preconditioning.
Other avenues for mitigating the computational complexity involve using structured automatic differentiation \citep{Ament:SBOSDA:2022}. 
Instead of using the gradient for taking stochastic steps, \citet{Penubothula:FirstOrder:2021} aim to find local critical points by querying where the predicted gradient is zero.




As mentioned above, fitting a Gaussian process to the objective allows for predicting gradients without having seen them {\it a priori} \cite[Sec 9.4]{RasmussenWilliams:GPs:2006}; \citet{Muller:LPSBO:2021} propose \textit{Gradient Information with BO} (GIBO), in which they guide local policy search in reinforcement learning tasks, exploiting this property. 
\citet{Nguyen:MPD:2022} address that expected gradients may not lead to the best performing outputs and compute the \textit{most probable descent direction}.

\subsection{Structured spaces}
\label{sec:taxonomy:structured-spaces}

Some applications work over structured spaces. For example, the angles of robot arms and protein backbones map to Riemannian manifolds \citep{Jaquier:GABO:2020, Penner:ProtMODULI:2022}, and input spaces might also contain mixed variables (i.e. products of real and categorical spaces). 
%
To compute non-linear embeddings (see Sec.~\ref{sec:taxonomy:non-linear-embeddings}) followed by standard Bayesian optimization (or small variations thereof) can allow us to work over such spaces. \citet{Jaquier:GABO:2020} use kernels defined on Riemannian manifolds \citep{Feragen:GeodesicKernels:2015,Borovitsky:MaternKernelOnManifold:2020} and optimize the acquisition function using tools from Riemannian optimization \citep{Boumal:ManifoldOpt:2023}. 
The authors expand their framework to high-dimensional manifolds by projecting to lower-dimensional submanifolds, which is roughly the equivalent to \textit{linear embeddings} in the Riemannian settings \citep{Jaquier:HDGABO:2020}. 

In the categorical and mixed-variable setting, kernels over string spaces \citep{Lodhi:SSKs:2000, Shervashidze:WLGraphKernel:2011}, can be applied to BO  \citep{Moss:BOSS:2020}. Other methods construct combinatorial graph and diffusion kernels-based GPs \citep{Oh:COMBO:2019}. 
\citet{Deshwal:LADDER:2021} combine latent space kernels with combinatorial kernels in an autoencoder-based approach. 

Recently, \citet{Daulton:PR:2022} have proposed a continuous relaxation of the discrete variables to ease the optimization of the acquisition function. \citet{Deshwal:BODI:2023} propose another way to map discrete variables to continuous space, relying on Hamming distances to make a dictionary for embeddings.
\citet{Papenmeier:BOUNCE:2024} extend previous work to both continuous and categorical variables: \texttt{BAxUS} learns an increasing sequence of subspaces using hash matrices which, when combined with the \texttt{CoCaBo} kernel \citep{Ru:CoCaBO:2020}, renders an algorithm for the mixed-variable setting.
Finally, through a continuous relaxation of the objective that incorporates {\it prior} pretrained models, \citet{Michael:COREL:2024} propose a surrogate on the probability vector space to optimize either the discrete input space or a continuous latent one. 

\paragraph{Other.} Some methods evade our taxonomy but are worth mentioning: some focus on the optimization of the acquisition function and the impact of initializations \citep{Zhao:AIBO:2024,Ngo:CMABO:2024}. Other methods balance both active learning (i.e.\@ building a better surrogate model) and optimization \citep{Hvarfner:SCOREBO:2023}.
Most recently, two articles claimed that the standard setting for Bayesian optimization or slight variations of it perform as well as the state-of-the-art of all the aforementioned families \citep{Hvarfner:VanilaBO:2024, Xu:StdVanillaBO:2024} -- begging the question, can these methods optimize in high dimensional discrete problem spaces in a sample efficient manner?

\section{Benchmarking the performance of HDBO methods}


\begin{figure}
    \centering
    \includegraphics[width=1.0\columnwidth]{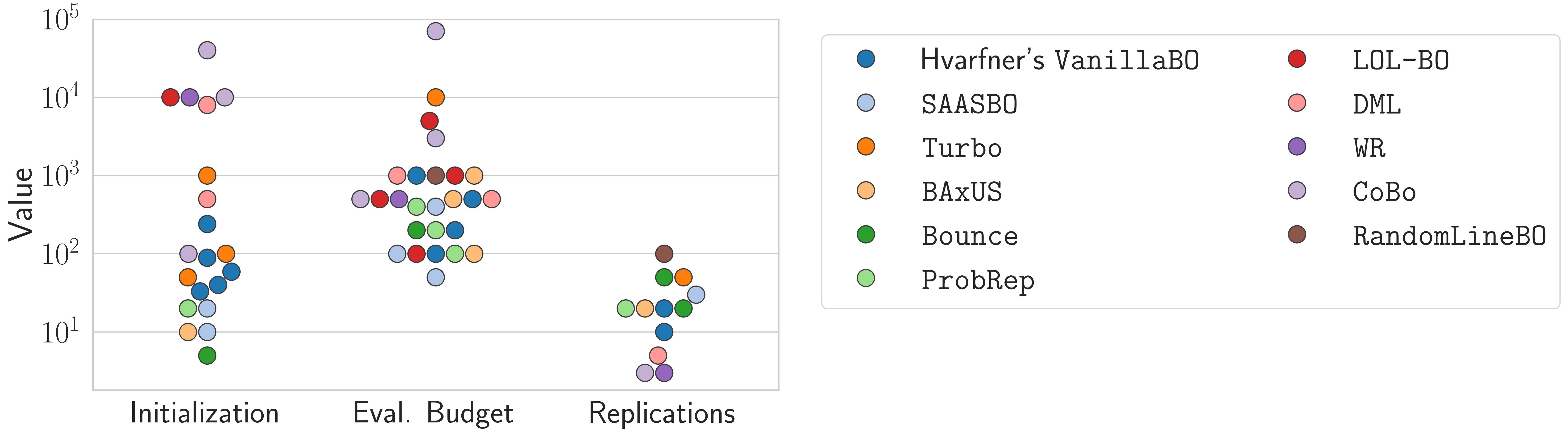}
    \vspace{-3mm}
    \caption{Initialization, evaluation budget, and nr.\@ of replications using different seeds reported in the experimental set-ups of several HDBO methods. We see heterogeneity in the evaluation of optimizers.}
    \label{fig:results:different-experimental-setups}
    \vspace{-1em}
\end{figure}

Practitioners that decide on what Bayesian optimization algorithm to use for their application will face several challenges.
While surveying the field, we noticed two key discrepancies in the reported experimental set-ups: (i) the initialization varies from as low as \emph{none} randomly/SOBOL sampled points to over $10^3$, (ii) evaluation budgets also vary for the same types of tasks.
Fig.~\ref{fig:results:different-experimental-setups} visualizes these different experimental set-ups as swarmplots.
Moreover, our survey covered code availability. The state-of-the-art is being pulled by workhorses, which have democratized access to GP and BO implementations: \texttt{GPyTorch} \citep{Gardner:Gpytorch:2018} and \texttt{BoTorch} \citep{Balandat:BoTorch:2020}, and \texttt{GPFlow} \citep{Matthews2017gpflow,Wilk2020gpflow} and \texttt{Trieste} \citep{Picheny2023trieste}. 
These libraries are highly useful and impactful, yet one can obtain cross-dependency conflicts between them especially if third-party dependencies are introduced or if very specific versions are required for solver setups. 
As a particular example, solvers like \texttt{ProbRep} cannot co-exist with \texttt{Ax}-based solvers like \texttt{SAASBO} or Hvarfner's \texttt{VanillaBO}. There is a need for \textit{isolating} optimizers, specifying up-to-date environments in which they can run. 
These issues led to the development of \poli{}.

\subsection{\poli{} and \texttt{poli-baselines}: a framework for benchmarking discrete optimizers}
\begin{wrapfigure}{r}{.52\textwidth}
\vspace{-0.55cm}
\begin{lstlisting}[language=python]
# pip install poli-core[ehrlich]
# pip install git+https://https://github.com/MachineLearningLifeScience/poli-baselines.git
from poli.repository import (
    EhrlichHoloProblemFactory,
)
from poli_baselines.solvers.simple.random_mutation import (
    RandomMutation
)

problem = EhrlichHoloProblemFactory().create(
    sequence_length=15,
    n_motifs=2,
    motif_length=7,
)
f, x0 = problem.black_box, problem.x0

solver = RandomMutation(
    black_box=f,
    x0=x0,
    y0=f(x0),
    greedy=True,  # Directed Evolution.
)

solver.solve(100)
\end{lstlisting}
\vspace{-1em}
\end{wrapfigure}
We want to solve truly high dimensional problems that are relevant for domains like biology and chemistry.
To make the outcomes comparable, we require a unified way of defining the problem which includes consistent starting points, budgets, runtime environments, relevant assets (i.e.\@models used for the black box), and a logging backend invoked for every oracle observation.
To that end, we implement the \textit{Protein Objectives Library} (\poli{})~ to provide potentially isolated black box functions. Building on open source tools, \poli{}~currently provides 35 black box tasks; besides the Practical Molecular Optimization (PMO) benchmark \citep{Huang:TDC:2021,Gao:PMOMolOpt:2022,Brown:Guacamol:2019}, it includes \texttt{Dockstring} \citep{Garcia:DOCKSTRING:2022} as well as other protein-related black boxes like stability and solvent accessibility \citep{Delgado:FOLDX5:2019, Blaabjerg:RASP:2023, Chapman:Biopython:2000, Stanton:LAMBO:2022}.\footnote{A complete list can be found in our documentation: \url{https://machinelearninglifescience.github.io/poli-docs/}}
The majority of black box functions can be queried with any string that complies with the corresponding alphabet making the oracles available for free-form optimization. This is an important distinction compared to pre-existing benchmarks that rely on pools of precompiled observations \citep{Notin:ProtGym:2023}.

We further provide an interface for the solvers used for the individual optimization tasks: \texttt{poli-baselines}.
Consistent, stable (and up to date) environments of individual optimizers can be found therein, as well as a standardized way to query them and solve the problems raised in the previous section. These environments and optimizers are tested weekly through GitHub actions, guaranteeing their usability.
An example of how \poli{} and \polibaselines{} work is provided above.
A problem containing a black box, an initial input, and potentially a data package is created through problem factories/a \texttt{create} method. A solver takes as input a black box and optionally supervised data, and uses the \texttt{solve} method to run the optimization for a given number of iterations. Such an interface can accommodate any optimizer that provides a method for running the optimization for a given amount of steps, as well as optimizers that do not accept custom initializations (e.g. \texttt{Bounce} or \texttt{BAxUS}).  Sec.~\ref{sec:appendix:technical-details-on-poli-and-poli-baselines} provides an introduction to this software's technical details, including a description of the logging logic though observers that keep track of every black box call.

\begin{table}[]
    \centering
    \resizebox{\textwidth}{!}{\begin{tabular}{llllll}
\toprule
Solver\textbackslash Oracle & PestControlEquiv & Ehrlich(L=5) & Ehrlich(L=15) & Ehrlich(L=64) & Sum (normalized per row) \\
\midrule
\texttt{DirectedEvolution} & \cellcolor{Green!47}$0.968{\pm \scriptstyle 0.03}$ & \cellcolor{Green!50}$1.000{\pm \scriptstyle 0.00}$ & \cellcolor{Green!12}$0.448{\pm \scriptstyle 0.16}$ & \cellcolor{Green!2}$0.114{\pm \scriptstyle 0.07}$ & \cellcolor{Green!50}$2.23{\pm \scriptstyle 0.27}$ \\
\texttt{HillClimbing} & \cellcolor{Green!19}$0.640{\pm \scriptstyle 0.12}$ & \cellcolor{Green!0}$0.500{\pm \scriptstyle 0.25}$ & \cellcolor{Green!8}$0.392{\pm \scriptstyle 0.20}$ & \cellcolor{Green!0}$0.089{\pm \scriptstyle 0.07}$ & \cellcolor{Green!0}$0.56{\pm \scriptstyle 0.64}$ \\
\texttt{CMAES} & \cellcolor{Green!34}$0.816{\pm \scriptstyle 0.06}$ & \cellcolor{Green!25}$0.750{\pm \scriptstyle 0.18}$ & \cellcolor{Green!2}$0.312{\pm \scriptstyle 0.13}$ & \cellcolor{Green!0}$0.077{\pm \scriptstyle 0.08}$ & \cellcolor{Green!20}$1.24{\pm \scriptstyle 0.45}$ \\
\texttt{GeneticAlgorithm} & \cellcolor{Green!25}$0.712{\pm \scriptstyle 0.02}$ & \cellcolor{Green!44}$0.950{\pm \scriptstyle 0.11}$ & \cellcolor{Green!4}$0.336{\pm \scriptstyle 0.10}$ & \cellcolor{Green!0}$0.083{\pm \scriptstyle 0.08}$ & \cellcolor{Green!28}$1.50{\pm \scriptstyle 0.31}$ \\
\midrule
Hvarfner's \texttt{VanillaBO} & \cellcolor{Green!43}$0.928{\pm \scriptstyle 0.08}$ & \cellcolor{Green!15}$0.650{\pm \scriptstyle 0.14}$ & \cellcolor{Green!3}$0.328{\pm \scriptstyle 0.18}$ & \cellcolor{Green!0}$0.079{\pm \scriptstyle 0.08}$ & \cellcolor{Green!20}$1.26{\pm \scriptstyle 0.47}$ \\
\texttt{RandomLineBO} & \cellcolor{Green!17}$0.624{\pm \scriptstyle 0.11}$ & \cellcolor{Green!19}$0.700{\pm \scriptstyle 0.27}$ & \cellcolor{Green!13}$0.472{\pm \scriptstyle 0.14}$ & \cellcolor{Green!0}$0.084{\pm \scriptstyle 0.08}$ & \cellcolor{Green!14}$1.04{\pm \scriptstyle 0.60}$ \\
\texttt{SAASBO} & \cellcolor{Green!32}$0.792{\pm \scriptstyle 0.05}$ & \cellcolor{Green!9}$0.600{\pm \scriptstyle 0.14}$ & \cellcolor{Green!3}$0.328{\pm \scriptstyle 0.21}$ & \cellcolor{Green!0}$0.075{\pm \scriptstyle 0.06}$ & \cellcolor{Green!10}$0.92{\pm \scriptstyle 0.46}$ \\
\texttt{Turbo} & \cellcolor{Green!41}$0.896{\pm \scriptstyle 0.04}$ & \cellcolor{Green!35}$0.850{\pm \scriptstyle 0.14}$ & \cellcolor{Green!14}$0.480{\pm \scriptstyle 0.12}$ & \cellcolor{Green!2}$0.124{\pm \scriptstyle 0.11}$ & \cellcolor{Green!38}$1.86{\pm \scriptstyle 0.40}$ \\
\texttt{BAxUS} & \cellcolor{Green!25}$0.712{\pm \scriptstyle 0.08}$ & \cellcolor{Green!5}$0.550{\pm \scriptstyle 0.11}$ & \cellcolor{Green!8}$0.400{\pm \scriptstyle 0.11}$ & \cellcolor{Green!0}$0.077{\pm \scriptstyle 0.08}$ & \cellcolor{Green!6}$0.79{\pm \scriptstyle 0.39}$ \\
\midrule
\texttt{Bounce} & \cellcolor{Green!50}$1.000{\pm \scriptstyle 0.00}$ & \cellcolor{Green!40}$0.900{\pm \scriptstyle 0.14}$ & \cellcolor{Green!9}$0.416{\pm \scriptstyle 0.13}$ & \cellcolor{Green!0}$0.076{\pm \scriptstyle 0.06}$ & \cellcolor{Green!43}$2.00{\pm \scriptstyle 0.33}$ \\
\texttt{ProbRep} & \cellcolor{Green!41}$0.896{\pm \scriptstyle 0.04}$ & \cellcolor{Green!44}$0.950{\pm \scriptstyle 0.11}$ & \cellcolor{Green!3}$0.328{\pm \scriptstyle 0.08}$ & \cellcolor{Green!0}$0.076{\pm \scriptstyle 0.08}$ & \cellcolor{Green!37}$1.80{\pm \scriptstyle 0.31}$ \\
\bottomrule
\end{tabular}
}
    \caption{Sequence design problems using Ehrlich functions. Solver's performance (measured as the average best value achieved during an optimization campaign of 300 iterations) is colored according to their closeness to the known optimal value (1.0). All problems except for Ehrlich with sequence length 64 were initialized with 10 supervised samples. The remaining one was initialized with 1000.}
    \label{tab:results:ehrlich}
    \vspace{-0.5cm}
\end{table}



\subsection{Benchmarking HDBO in discrete sequences}

Using the black boxes provided in \poli{}, as well as the solvers provided in \polibaselines{}, we benchmark the performance of high-dimensional Bayesian optimization solvers on discrete sequences. Such optimization can take place either at the sequence level (as solvers in the \textit{Structured Spaces} do), or in a continuous version via either one-hot representations or learned latent spaces. Benchmarks on both fronts are presented in this paper: from sequence design tasks of varying complexity (mimicking protein engineering) in one-hot/sequence space using Ehrlich functions \citep{Stanton:Ehrlich:2024}, to latent space optimization of small molecules on the Practical Molecular Optimization (PMO) benchmark \citep{Gao:PMOMolOpt:2022,Huang:TDC:2021,Brown:Guacamol:2019}.\footnote{We emphasize that the results provided in this paper are continuously updated on \href{https://machinelearninglifescience.github.io/hdbo_benchmark/}{our project's website}.}

\subsubsection{Sequence design problems of varying complexity}
\label{sec:benchmark:ehrlich}

Ehrlich functions \citep{Stanton:Ehrlich:2024} are closed-form procedurally-generated oracles in which a certain collection of motifs needs to be satisfied in a sequence of a pre-selected length. The oracle's score is between 0 and 1, and determined by how much of the individual motifs are satisfied. The number of motifs and their length are hyperparameters specified by the user. Sec.~\ref{sec:appendix:ehrlich} provides a detailed introduction.
In particular, we test on 4 different configurations: a PestControl equivalent\footnote{PestControl is a black box whose input are 25 categorical variables, each of which has 5 potential categories. It is a commonly tested black box in the \textit{Structured Spaces} literature.} with an alphabet size of 5 and a sequence length of 25 (i.e.\ one motif of length 25). Further, 3 different configurations imitating protein design using the alphabet of 20 amino acids and sequence lengths of 5, 15, and 64 are tested. The motif lengths and number of motifs are (4, 1), (7, 2) and (10, 4) respectively. We initialize PestControlEquiv, and the two small Ehrlich problems with 10 supervised samples, and the latter with 1000, and all methods have an evaluation budget of 300.

We optimize these oracles with representatives from the taxonomy that are frequently tested in the HDBO literature. We select Hvarfner's \texttt{VanillaBO}, \texttt{RandomLineBO}, \texttt{Turbo}, \texttt{BAxUS}, \texttt{SAASBO}, \texttt{Bounce}, and \texttt{ProbRep}, including also baselines like \texttt{DirectedEvolution} (i.e. greedily mutating the incumbent in sequence space), \texttt{HillClimbing} (which explores the input space by taking random Gaussian steps), and evolutionary algorithms/strategies like \texttt{GeneticAlgorithm} and \texttt{CMA-ES}. Notice that, due to the nature of their implementations, \texttt{BAxUS} and \texttt{Bounce} are not initialized with the same supervised data (but rather according to their implementation: SOBOL sampling in a linear subspace). In this scenario, the continuous solvers are optimizing over one-hot inputs, and the discrete ones work over sequence space. The next section explores latent space optimization.

Table~\ref{tab:results:ehrlich} shows the mean best value achieved in the aforementioned 4 problems with one standard deviation for 5 seeds. Problems of low complexity (i.e. PestControlEquiv and Ehrlich with seq. length 5) are readily solvable using one-hot BO, with \texttt{Bounce} solving PestControlEquiv in all 5 replications. Such problems are also easily solvable by naïve baselines like \texttt{DirectedEvolution}. For more complex problems, BO optimizers perform equally or worse than greedy baselines. We hypothesize that naïve baselines will perform worse on black boxes with higher-order effects. Note that \texttt{SAASBO} and \texttt{ProbRep} ran out of memory when instantiating the largest problem.

\subsubsection{Benchmarking HDBO on PMO}
\label{sec:benchmark:pmo}

\begin{table}[]
    \centering
    \resizebox{\textwidth}{!}{\begin{tabular}{l|lll|lllll|ll}
\toprule
 & \texttt{Hill} & \texttt{Genetic} & \texttt{CMAES} & Hvarfner's & \texttt{Random} & \texttt{SAASBO} & \texttt{BAxUS} & \texttt{Turbo} & \texttt{Bounce} & \texttt{ProbRep} \\
Oracle & \texttt{Climbing} & \texttt{Algorithm}  &  & \texttt{VanillaBO} & \texttt{LineBO} &  &  &  &  &  \\
\midrule
albuterol\_similarity & \cellcolor{Green!7}$0.20{\pm \scriptstyle 0.06}$ & \cellcolor{Green!29}$0.32{\pm \scriptstyle 0.03}$ & \cellcolor{Green!50}$0.44{\pm \scriptstyle 0.03}$ & \cellcolor{Green!5}$0.19{\pm \scriptstyle 0.01}$ & \cellcolor{Green!0}$0.15{\pm \scriptstyle 0.04}$ & \cellcolor{Green!48}$0.43{\pm \scriptstyle 0.05}$ & \cellcolor{Green!31}$0.33{\pm \scriptstyle 0.09}$ & \cellcolor{Green!33}$0.35{\pm \scriptstyle 0.03}$ & \cellcolor{Green!0}$0.16{\pm \scriptstyle 0.01}$ & \cellcolor{Green!9}$0.21{\pm \scriptstyle 0.03}$ \\
amlodipine\_mpo & \cellcolor{Green!18}$0.16{\pm \scriptstyle 0.07}$ & \cellcolor{Green!35}$0.29{\pm \scriptstyle 0.06}$ & \cellcolor{Green!49}$0.41{\pm \scriptstyle 0.01}$ & \cellcolor{Green!14}$0.12{\pm \scriptstyle 0.07}$ & \cellcolor{Green!2}$0.02{\pm \scriptstyle 0.03}$ & \cellcolor{Green!49}$0.41{\pm \scriptstyle 0.02}$ & \cellcolor{Green!49}$0.41{\pm \scriptstyle 0.02}$ & \cellcolor{Green!50}$0.42{\pm \scriptstyle 0.03}$ & \cellcolor{Green!0}$0.00{\pm \scriptstyle 0.00}$ & \cellcolor{Green!0}$0.00{\pm \scriptstyle 0.00}$ \\
celecoxib\_rediscovery & \cellcolor{Green!8}$0.04{\pm \scriptstyle 0.02}$ & \cellcolor{Green!20}$0.08{\pm \scriptstyle 0.00}$ & \cellcolor{Green!43}$0.16{\pm \scriptstyle 0.01}$ & \cellcolor{Green!10}$0.05{\pm \scriptstyle 0.01}$ & \cellcolor{Green!4}$0.02{\pm \scriptstyle 0.00}$ & \cellcolor{Green!50}$0.18{\pm \scriptstyle 0.04}$ & \cellcolor{Green!43}$0.16{\pm \scriptstyle 0.01}$ & \cellcolor{Green!43}$0.16{\pm \scriptstyle 0.05}$ & \cellcolor{Green!4}$0.02{\pm \scriptstyle 0.01}$ & \cellcolor{Green!2}$0.02{\pm \scriptstyle 0.00}$ \\
deco\_hop & \cellcolor{Green!28}$0.53{\pm \scriptstyle 0.01}$ & \cellcolor{Green!21}$0.52{\pm \scriptstyle 0.00}$ & \cellcolor{Green!50}$0.54{\pm \scriptstyle 0.01}$ & \cellcolor{Green!33}$0.53{\pm \scriptstyle 0.00}$ & \cellcolor{Green!20}$0.52{\pm \scriptstyle 0.00}$ & \cellcolor{Green!40}$0.53{\pm \scriptstyle 0.01}$ & \cellcolor{Green!39}$0.53{\pm \scriptstyle 0.02}$ & \cellcolor{Green!36}$0.53{\pm \scriptstyle 0.02}$ & \cellcolor{Green!0}$0.50{\pm \scriptstyle 0.00}$ & \cellcolor{Green!6}$0.51{\pm \scriptstyle 0.00}$ \\
drd2\_docking & \cellcolor{Green!44}$0.03{\pm \scriptstyle 0.00}$ & \cellcolor{Green!44}$0.03{\pm \scriptstyle 0.00}$ & \cellcolor{Green!46}$0.03{\pm \scriptstyle 0.00}$ & \cellcolor{Green!44}$0.03{\pm \scriptstyle 0.00}$ & \cellcolor{Green!12}$0.02{\pm \scriptstyle 0.01}$ & \cellcolor{Green!50}$0.03{\pm \scriptstyle 0.00}$ & \cellcolor{Green!44}$0.03{\pm \scriptstyle 0.00}$ & \cellcolor{Green!44}$0.03{\pm \scriptstyle 0.00}$ & \cellcolor{Green!0}$0.01{\pm \scriptstyle 0.00}$ & \cellcolor{Green!4}$0.01{\pm \scriptstyle 0.00}$ \\
fexofenadine\_mpo & \cellcolor{Green!22}$0.37{\pm \scriptstyle 0.15}$ & \cellcolor{Green!32}$0.48{\pm \scriptstyle 0.03}$ & \cellcolor{Green!50}$0.66{\pm \scriptstyle 0.01}$ & \cellcolor{Green!25}$0.40{\pm \scriptstyle 0.03}$ & \cellcolor{Green!13}$0.27{\pm \scriptstyle 0.02}$ & \cellcolor{Green!47}$0.64{\pm \scriptstyle 0.04}$ & \cellcolor{Green!33}$0.49{\pm \scriptstyle 0.30}$ & \cellcolor{Green!37}$0.54{\pm \scriptstyle 0.24}$ & \cellcolor{Green!0}$0.13{\pm \scriptstyle 0.13}$ & \cellcolor{Green!6}$0.20{\pm \scriptstyle 0.08}$ \\
gsk3\_beta & \cellcolor{Green!37}$0.23{\pm \scriptstyle 0.11}$ & \cellcolor{Green!28}$0.19{\pm \scriptstyle 0.01}$ & \cellcolor{Green!29}$0.20{\pm \scriptstyle 0.04}$ & \cellcolor{Green!50}$0.27{\pm \scriptstyle 0.03}$ & \cellcolor{Green!29}$0.20{\pm \scriptstyle 0.07}$ & \cellcolor{Green!16}$0.15{\pm \scriptstyle 0.03}$ & \cellcolor{Green!7}$0.12{\pm \scriptstyle 0.04}$ & \cellcolor{Green!11}$0.13{\pm \scriptstyle 0.05}$ & \cellcolor{Green!0}$0.09{\pm \scriptstyle 0.08}$ & \cellcolor{Green!9}$0.12{\pm \scriptstyle 0.02}$ \\
isomer\_c7h8n2o2 & \cellcolor{Green!42}$0.63{\pm \scriptstyle 0.15}$ & \cellcolor{Green!49}$0.72{\pm \scriptstyle 0.09}$ & \cellcolor{Green!50}$0.73{\pm \scriptstyle 0.12}$ & \cellcolor{Green!33}$0.49{\pm \scriptstyle 0.04}$ & \cellcolor{Green!0}$0.03{\pm \scriptstyle 0.02}$ & \cellcolor{Green!44}$0.66{\pm \scriptstyle 0.18}$ & \cellcolor{Green!32}$0.49{\pm \scriptstyle 0.05}$ & \cellcolor{Green!12}$0.20{\pm \scriptstyle 0.17}$ & \cellcolor{Green!5}$0.11{\pm \scriptstyle 0.09}$ & \cellcolor{Green!15}$0.24{\pm \scriptstyle 0.11}$ \\
isomer\_c9h10n2o2pf2cl & \cellcolor{Green!42}$0.48{\pm \scriptstyle 0.16}$ & \cellcolor{Green!49}$0.56{\pm \scriptstyle 0.05}$ & \cellcolor{Green!50}$0.57{\pm \scriptstyle 0.14}$ & \cellcolor{Green!49}$0.56{\pm \scriptstyle 0.05}$ & \cellcolor{Green!28}$0.32{\pm \scriptstyle 0.21}$ & \cellcolor{Green!37}$0.43{\pm \scriptstyle 0.19}$ & \cellcolor{Green!31}$0.36{\pm \scriptstyle 0.16}$ & \cellcolor{Green!32}$0.37{\pm \scriptstyle 0.22}$ & \cellcolor{Green!0}$0.01{\pm \scriptstyle 0.01}$ & \cellcolor{Green!4}$0.06{\pm \scriptstyle 0.03}$ \\
jnk3 & \cellcolor{Green!50}$0.19{\pm \scriptstyle 0.03}$ & \cellcolor{Green!18}$0.10{\pm \scriptstyle 0.01}$ & \cellcolor{Green!19}$0.10{\pm \scriptstyle 0.03}$ & \cellcolor{Green!39}$0.16{\pm \scriptstyle 0.04}$ & \cellcolor{Green!19}$0.10{\pm \scriptstyle 0.03}$ & \cellcolor{Green!10}$0.08{\pm \scriptstyle 0.02}$ & \cellcolor{Green!10}$0.08{\pm \scriptstyle 0.02}$ & \cellcolor{Green!14}$0.09{\pm \scriptstyle 0.05}$ & \cellcolor{Green!0}$0.05{\pm \scriptstyle 0.04}$ & \cellcolor{Green!3}$0.06{\pm \scriptstyle 0.01}$ \\
median\_1 & \cellcolor{Green!8}$0.05{\pm \scriptstyle 0.03}$ & \cellcolor{Green!50}$0.18{\pm \scriptstyle 0.03}$ & \cellcolor{Green!45}$0.17{\pm \scriptstyle 0.01}$ & \cellcolor{Green!9}$0.05{\pm \scriptstyle 0.01}$ & \cellcolor{Green!0}$0.02{\pm \scriptstyle 0.02}$ & \cellcolor{Green!44}$0.16{\pm \scriptstyle 0.01}$ & \cellcolor{Green!31}$0.12{\pm \scriptstyle 0.01}$ & \cellcolor{Green!36}$0.14{\pm \scriptstyle 0.02}$ & \cellcolor{Green!3}$0.03{\pm \scriptstyle 0.01}$ & \cellcolor{Green!2}$0.02{\pm \scriptstyle 0.00}$ \\
median\_2 & \cellcolor{Green!2}$0.02{\pm \scriptstyle 0.01}$ & \cellcolor{Green!30}$0.08{\pm \scriptstyle 0.01}$ & \cellcolor{Green!50}$0.12{\pm \scriptstyle 0.00}$ & \cellcolor{Green!8}$0.03{\pm \scriptstyle 0.01}$ & \cellcolor{Green!0}$0.01{\pm \scriptstyle 0.00}$ & \cellcolor{Green!49}$0.12{\pm \scriptstyle 0.00}$ & \cellcolor{Green!49}$0.12{\pm \scriptstyle 0.02}$ & \cellcolor{Green!49}$0.12{\pm \scriptstyle 0.02}$ & \cellcolor{Green!1}$0.01{\pm \scriptstyle 0.00}$ & \cellcolor{Green!0}$0.01{\pm \scriptstyle 0.00}$ \\
mestranol\_similarity & \cellcolor{Green!12}$0.10{\pm \scriptstyle 0.09}$ & \cellcolor{Green!34}$0.26{\pm \scriptstyle 0.00}$ & \cellcolor{Green!50}$0.38{\pm \scriptstyle 0.02}$ & \cellcolor{Green!22}$0.18{\pm \scriptstyle 0.03}$ & \cellcolor{Green!7}$0.06{\pm \scriptstyle 0.03}$ & \cellcolor{Green!46}$0.35{\pm \scriptstyle 0.05}$ & \cellcolor{Green!44}$0.34{\pm \scriptstyle 0.05}$ & \cellcolor{Green!39}$0.30{\pm \scriptstyle 0.02}$ & \cellcolor{Green!0}$0.01{\pm \scriptstyle 0.00}$ & \cellcolor{Green!0}$0.02{\pm \scriptstyle 0.00}$ \\
osimetrinib\_mpo & \cellcolor{Green!48}$0.62{\pm \scriptstyle 0.01}$ & \cellcolor{Green!50}$0.63{\pm \scriptstyle 0.01}$ & \cellcolor{Green!35}$0.51{\pm \scriptstyle 0.30}$ & \cellcolor{Green!46}$0.60{\pm \scriptstyle 0.01}$ & \cellcolor{Green!44}$0.59{\pm \scriptstyle 0.01}$ & \cellcolor{Green!44}$0.59{\pm \scriptstyle 0.06}$ & \cellcolor{Green!0}$0.22{\pm \scriptstyle 0.35}$ & \cellcolor{Green!14}$0.33{\pm \scriptstyle 0.32}$ & \cellcolor{Green!9}$0.30{\pm \scriptstyle 0.31}$ & \cellcolor{Green!44}$0.59{\pm \scriptstyle 0.04}$ \\
perindopril\_mpo & \cellcolor{Green!0}$0.00{\pm \scriptstyle 0.00}$ & \cellcolor{Green!18}$0.10{\pm \scriptstyle 0.03}$ & \cellcolor{Green!49}$0.25{\pm \scriptstyle 0.10}$ & \cellcolor{Green!0}$0.00{\pm \scriptstyle 0.00}$ & \cellcolor{Green!3}$0.02{\pm \scriptstyle 0.03}$ & \cellcolor{Green!50}$0.25{\pm \scriptstyle 0.08}$ & \cellcolor{Green!47}$0.24{\pm \scriptstyle 0.14}$ & \cellcolor{Green!43}$0.22{\pm \scriptstyle 0.13}$ & \cellcolor{Green!0}$0.00{\pm \scriptstyle 0.00}$ & \cellcolor{Green!0}$0.00{\pm \scriptstyle 0.00}$ \\
ranolazine\_mpo & \cellcolor{Green!24}$0.31{\pm \scriptstyle 0.22}$ & \cellcolor{Green!41}$0.53{\pm \scriptstyle 0.02}$ & \cellcolor{Green!47}$0.59{\pm \scriptstyle 0.03}$ & \cellcolor{Green!20}$0.26{\pm \scriptstyle 0.15}$ & \cellcolor{Green!5}$0.07{\pm \scriptstyle 0.03}$ & \cellcolor{Green!50}$0.63{\pm \scriptstyle 0.01}$ & \cellcolor{Green!42}$0.54{\pm \scriptstyle 0.17}$ & \cellcolor{Green!38}$0.48{\pm \scriptstyle 0.19}$ & \cellcolor{Green!0}$0.00{\pm \scriptstyle 0.00}$ & \cellcolor{Green!8}$0.11{\pm \scriptstyle 0.02}$ \\
rdkit\_logp & \cellcolor{Green!5}$5.00{\pm \scriptstyle 1.85}$ & \cellcolor{Green!27}$13.28{\pm \scriptstyle 0.41}$ & \cellcolor{Green!50}$21.62{\pm \scriptstyle 0.12}$ & \cellcolor{Green!7}$5.93{\pm \scriptstyle 1.19}$ & \cellcolor{Green!6}$5.60{\pm \scriptstyle 4.35}$ & \cellcolor{Green!45}$19.87{\pm \scriptstyle 1.21}$ & \cellcolor{Green!39}$17.84{\pm \scriptstyle 2.53}$ & \cellcolor{Green!47}$20.87{\pm \scriptstyle 1.85}$ & \cellcolor{Green!0}$3.12{\pm \scriptstyle 1.20}$ & \cellcolor{Green!6}$5.49{\pm \scriptstyle 3.01}$ \\
rdkit\_qed & \cellcolor{Green!12}$0.55{\pm \scriptstyle 0.03}$ & \cellcolor{Green!25}$0.66{\pm \scriptstyle 0.07}$ & \cellcolor{Green!50}$0.90{\pm \scriptstyle 0.04}$ & \cellcolor{Green!19}$0.61{\pm \scriptstyle 0.02}$ & \cellcolor{Green!0}$0.42{\pm \scriptstyle 0.03}$ & \cellcolor{Green!39}$0.79{\pm \scriptstyle 0.13}$ & \cellcolor{Green!39}$0.80{\pm \scriptstyle 0.09}$ & \cellcolor{Green!34}$0.75{\pm \scriptstyle 0.13}$ & \cellcolor{Green!10}$0.52{\pm \scriptstyle 0.09}$ & \cellcolor{Green!18}$0.60{\pm \scriptstyle 0.05}$ \\
sa\_tdc & \cellcolor{Green!49}$8.70{\pm \scriptstyle 0.22}$ & \cellcolor{Green!50}$8.71{\pm \scriptstyle 0.20}$ & \cellcolor{Green!30}$7.48{\pm \scriptstyle 0.31}$ & \cellcolor{Green!49}$8.69{\pm \scriptstyle 0.10}$ & \cellcolor{Green!41}$8.17{\pm \scriptstyle 0.83}$ & \cellcolor{Green!25}$7.14{\pm \scriptstyle 0.88}$ & \cellcolor{Green!31}$7.56{\pm \scriptstyle 0.05}$ & \cellcolor{Green!0}$5.55{\pm \scriptstyle 0.31}$ & \cellcolor{Green!44}$8.36{\pm \scriptstyle 0.46}$ & \cellcolor{Green!48}$8.59{\pm \scriptstyle 0.13}$ \\
scaffold\_hop & \cellcolor{Green!33}$0.37{\pm \scriptstyle 0.01}$ & \cellcolor{Green!26}$0.36{\pm \scriptstyle 0.00}$ & \cellcolor{Green!50}$0.39{\pm \scriptstyle 0.01}$ & \cellcolor{Green!40}$0.38{\pm \scriptstyle 0.01}$ & \cellcolor{Green!33}$0.37{\pm \scriptstyle 0.00}$ & \cellcolor{Green!47}$0.38{\pm \scriptstyle 0.01}$ & \cellcolor{Green!29}$0.37{\pm \scriptstyle 0.01}$ & \cellcolor{Green!33}$0.37{\pm \scriptstyle 0.02}$ & \cellcolor{Green!0}$0.34{\pm \scriptstyle 0.01}$ & \cellcolor{Green!3}$0.34{\pm \scriptstyle 0.00}$ \\
sitagliptin\_mpo & \cellcolor{Green!29}$0.10{\pm \scriptstyle 0.11}$ & \cellcolor{Green!15}$0.05{\pm \scriptstyle 0.05}$ & \cellcolor{Green!46}$0.15{\pm \scriptstyle 0.15}$ & \cellcolor{Green!37}$0.12{\pm \scriptstyle 0.13}$ & \cellcolor{Green!16}$0.05{\pm \scriptstyle 0.05}$ & \cellcolor{Green!32}$0.11{\pm \scriptstyle 0.12}$ & \cellcolor{Green!26}$0.09{\pm \scriptstyle 0.07}$ & \cellcolor{Green!50}$0.16{\pm \scriptstyle 0.06}$ & \cellcolor{Green!0}$0.00{\pm \scriptstyle 0.00}$ & \cellcolor{Green!0}$0.00{\pm \scriptstyle 0.00}$ \\
thiothixene\_rediscovery & \cellcolor{Green!3}$0.03{\pm \scriptstyle 0.02}$ & \cellcolor{Green!29}$0.13{\pm \scriptstyle 0.03}$ & \cellcolor{Green!45}$0.20{\pm \scriptstyle 0.05}$ & \cellcolor{Green!7}$0.05{\pm \scriptstyle 0.01}$ & \cellcolor{Green!0}$0.02{\pm \scriptstyle 0.00}$ & \cellcolor{Green!50}$0.22{\pm \scriptstyle 0.02}$ & \cellcolor{Green!44}$0.19{\pm \scriptstyle 0.04}$ & \cellcolor{Green!45}$0.20{\pm \scriptstyle 0.03}$ & \cellcolor{Green!0}$0.02{\pm \scriptstyle 0.01}$ & \cellcolor{Green!2}$0.03{\pm \scriptstyle 0.01}$ \\
troglitazone\_rediscovery & \cellcolor{Green!7}$0.04{\pm \scriptstyle 0.03}$ & \cellcolor{Green!32}$0.11{\pm \scriptstyle 0.02}$ & \cellcolor{Green!50}$0.16{\pm \scriptstyle 0.01}$ & \cellcolor{Green!10}$0.05{\pm \scriptstyle 0.01}$ & \cellcolor{Green!0}$0.02{\pm \scriptstyle 0.00}$ & \cellcolor{Green!48}$0.16{\pm \scriptstyle 0.01}$ & \cellcolor{Green!39}$0.13{\pm \scriptstyle 0.01}$ & \cellcolor{Green!46}$0.15{\pm \scriptstyle 0.01}$ & \cellcolor{Green!0}$0.02{\pm \scriptstyle 0.01}$ & \cellcolor{Green!0}$0.02{\pm \scriptstyle 0.00}$ \\
valsartan\_smarts & \cellcolor{Green!0}$0.00{\pm \scriptstyle 0.00}$ & \cellcolor{Green!0}$0.00{\pm \scriptstyle 0.00}$ & \cellcolor{Green!0}$0.00{\pm \scriptstyle 0.00}$ & \cellcolor{Green!0}$0.00{\pm \scriptstyle 0.00}$ & \cellcolor{Green!0}$0.00{\pm \scriptstyle 0.00}$ & \cellcolor{Green!0}$0.00{\pm \scriptstyle 0.00}$ & \cellcolor{Green!0}$0.00{\pm \scriptstyle 0.00}$ & \cellcolor{Green!0}$0.00{\pm \scriptstyle 0.00}$ & \cellcolor{Green!0}$0.00{\pm \scriptstyle 0.00}$ & \cellcolor{Green!0}$0.00{\pm \scriptstyle 0.00}$ \\
zaleplon\_mpo & \cellcolor{Green!7}$0.05{\pm \scriptstyle 0.09}$ & \cellcolor{Green!7}$0.05{\pm \scriptstyle 0.00}$ & \cellcolor{Green!19}$0.13{\pm \scriptstyle 0.04}$ & \cellcolor{Green!13}$0.09{\pm \scriptstyle 0.03}$ & \cellcolor{Green!0}$0.00{\pm \scriptstyle 0.00}$ & \cellcolor{Green!50}$0.33{\pm \scriptstyle 0.08}$ & \cellcolor{Green!8}$0.05{\pm \scriptstyle 0.05}$ & \cellcolor{Green!19}$0.13{\pm \scriptstyle 0.11}$ & \cellcolor{Green!0}$0.00{\pm \scriptstyle 0.00}$ & \cellcolor{Green!0}$0.00{\pm \scriptstyle 0.00}$ \\
\midrule
Sum (normalized per row) & \cellcolor{Green!23}$10.92{\pm \scriptstyle 3.46}$ & \cellcolor{Green!35}$15.39{\pm \scriptstyle 1.19}$ & \cellcolor{Green!50}$21.15{\pm \scriptstyle 1.57}$ & \cellcolor{Green!26}$11.99{\pm \scriptstyle 1.98}$ & \cellcolor{Green!10}$5.81{\pm \scriptstyle 5.82}$ & \cellcolor{Green!48}$20.42{\pm \scriptstyle 3.26}$ & \cellcolor{Green!36}$15.95{\pm \scriptstyle 4.28}$ & \cellcolor{Green!37}$16.30{\pm \scriptstyle 4.08}$ & \cellcolor{Green!0}$1.59{\pm \scriptstyle 2.47}$ & \cellcolor{Green!6}$3.95{\pm \scriptstyle 3.57}$ \\
\bottomrule
\end{tabular}
}
    \caption{Results on the PMO benchmark for a 128-latent space. The best output of the optimization campaign over max.\@ 300 iterations are averaged over three runs, using a Sobol-sampled initialization of 10 latent points. The last row is computed by adding the result of min-max normalizing each row. Note that \texttt{Bounce} consistently ran out of memory in as few iterations as 40 (where the dimensionality of the ongoing subspace is increased), and \texttt{ProbRep} runs were stopped after 24 hours.}
    \label{tab:results:absolute_values_for_128_latent_dim}
    \vspace{-0.5cm}
\end{table}

\begin{table}
\resizebox{\textwidth}{!}{\begin{tabular}{l|lll|llll|ll}
\toprule
 & \texttt{Hill} & \texttt{Genetic} & \texttt{CMAES} & Hvarfner's & \texttt{Random} & \texttt{SAASBO} & \texttt{Turbo} & \texttt{Bounce} & \texttt{ProbRep} \\
Oracle & \texttt{Climbing} & \texttt{Algorithm}  &  & \texttt{VanillaBO}  & \texttt{LineBO} &  &  &  &  \\
\midrule
albuterol\_similarity & \cellcolor{Green!31}$0.31{\pm \scriptstyle 0.10}$ & \cellcolor{Green!19}$0.26{\pm \scriptstyle 0.01}$ & \cellcolor{Green!45}$0.38{\pm \scriptstyle 0.05}$ & \cellcolor{Green!50}$0.40{\pm \scriptstyle 0.02}$ & \cellcolor{Green!41}$0.36{\pm \scriptstyle 0.02}$ & \cellcolor{Green!42}$0.36{\pm \scriptstyle 0.07}$ & \cellcolor{Green!49}$0.40{\pm \scriptstyle 0.08}$ & \cellcolor{Green!0}$0.17{\pm \scriptstyle 0.02}$ & \cellcolor{Green!7}$0.21{\pm \scriptstyle 0.03}$ \\
amlodipine\_mpo & \cellcolor{Green!37}$0.30{\pm \scriptstyle 0.04}$ & \cellcolor{Green!37}$0.30{\pm \scriptstyle 0.05}$ & \cellcolor{Green!44}$0.36{\pm \scriptstyle 0.04}$ & \cellcolor{Green!42}$0.34{\pm \scriptstyle 0.02}$ & \cellcolor{Green!50}$0.40{\pm \scriptstyle 0.05}$ & \cellcolor{Green!48}$0.39{\pm \scriptstyle 0.06}$ & \cellcolor{Green!46}$0.37{\pm \scriptstyle 0.03}$ & \cellcolor{Green!0}$0.00{\pm \scriptstyle 0.00}$ & \cellcolor{Green!0}$0.00{\pm \scriptstyle 0.00}$ \\
celecoxib\_rediscovery & \cellcolor{Green!28}$0.09{\pm \scriptstyle 0.01}$ & \cellcolor{Green!17}$0.06{\pm \scriptstyle 0.02}$ & \cellcolor{Green!40}$0.13{\pm \scriptstyle 0.04}$ & \cellcolor{Green!42}$0.13{\pm \scriptstyle 0.01}$ & \cellcolor{Green!44}$0.14{\pm \scriptstyle 0.00}$ & \cellcolor{Green!43}$0.14{\pm \scriptstyle 0.01}$ & \cellcolor{Green!42}$0.13{\pm \scriptstyle 0.04}$ & \cellcolor{Green!2}$0.02{\pm \scriptstyle 0.01}$ & \cellcolor{Green!0}$0.02{\pm \scriptstyle 0.00}$ \\
deco\_hop & \cellcolor{Green!5}$0.51{\pm \scriptstyle 0.00}$ & \cellcolor{Green!18}$0.52{\pm \scriptstyle 0.01}$ & \cellcolor{Green!33}$0.52{\pm \scriptstyle 0.01}$ & \cellcolor{Green!42}$0.53{\pm \scriptstyle 0.01}$ & \cellcolor{Green!50}$0.53{\pm \scriptstyle 0.01}$ & \cellcolor{Green!29}$0.52{\pm \scriptstyle 0.00}$ & \cellcolor{Green!49}$0.53{\pm \scriptstyle 0.01}$ & \cellcolor{Green!0}$0.50{\pm \scriptstyle 0.00}$ & \cellcolor{Green!7}$0.51{\pm \scriptstyle 0.00}$ \\
drd2\_docking & \cellcolor{Green!43}$0.03{\pm \scriptstyle 0.00}$ & \cellcolor{Green!40}$0.03{\pm \scriptstyle 0.00}$ & \cellcolor{Green!39}$0.03{\pm \scriptstyle 0.00}$ & \cellcolor{Green!50}$0.03{\pm \scriptstyle 0.01}$ & \cellcolor{Green!39}$0.03{\pm \scriptstyle 0.00}$ & \cellcolor{Green!17}$0.02{\pm \scriptstyle 0.01}$ & \cellcolor{Green!39}$0.03{\pm \scriptstyle 0.00}$ & \cellcolor{Green!0}$0.01{\pm \scriptstyle 0.00}$ & \cellcolor{Green!4}$0.01{\pm \scriptstyle 0.00}$ \\
fexofenadine\_mpo & \cellcolor{Green!0}$0.01{\pm \scriptstyle 0.01}$ & \cellcolor{Green!33}$0.43{\pm \scriptstyle 0.02}$ & \cellcolor{Green!44}$0.55{\pm \scriptstyle 0.01}$ & \cellcolor{Green!48}$0.60{\pm \scriptstyle 0.04}$ & \cellcolor{Green!45}$0.57{\pm \scriptstyle 0.02}$ & \cellcolor{Green!47}$0.59{\pm \scriptstyle 0.07}$ & \cellcolor{Green!17}$0.23{\pm \scriptstyle 0.30}$ & \cellcolor{Green!9}$0.13{\pm \scriptstyle 0.13}$ & \cellcolor{Green!15}$0.20{\pm \scriptstyle 0.08}$ \\
gsk3\_beta & \cellcolor{Green!0}$0.07{\pm \scriptstyle 0.02}$ & \cellcolor{Green!30}$0.19{\pm \scriptstyle 0.01}$ & \cellcolor{Green!22}$0.16{\pm \scriptstyle 0.01}$ & \cellcolor{Green!18}$0.14{\pm \scriptstyle 0.02}$ & \cellcolor{Green!32}$0.20{\pm \scriptstyle 0.02}$ & \cellcolor{Green!27}$0.18{\pm \scriptstyle 0.05}$ & \cellcolor{Green!5}$0.09{\pm \scriptstyle 0.03}$ & \cellcolor{Green!4}$0.09{\pm \scriptstyle 0.08}$ & \cellcolor{Green!13}$0.12{\pm \scriptstyle 0.02}$ \\
isomer\_c7h8n2o2 & \cellcolor{Green!23}$0.45{\pm \scriptstyle 0.05}$ & \cellcolor{Green!35}$0.63{\pm \scriptstyle 0.08}$ & \cellcolor{Green!21}$0.42{\pm \scriptstyle 0.10}$ & \cellcolor{Green!47}$0.81{\pm \scriptstyle 0.07}$ & \cellcolor{Green!46}$0.79{\pm \scriptstyle 0.10}$ & \cellcolor{Green!50}$0.84{\pm \scriptstyle 0.09}$ & \cellcolor{Green!9}$0.25{\pm \scriptstyle 0.13}$ & \cellcolor{Green!0}$0.11{\pm \scriptstyle 0.09}$ & \cellcolor{Green!9}$0.24{\pm \scriptstyle 0.11}$ \\
isomer\_c9h10n2o2pf2cl & \cellcolor{Green!26}$0.30{\pm \scriptstyle 0.14}$ & \cellcolor{Green!39}$0.46{\pm \scriptstyle 0.17}$ & \cellcolor{Green!35}$0.41{\pm \scriptstyle 0.20}$ & \cellcolor{Green!41}$0.49{\pm \scriptstyle 0.11}$ & \cellcolor{Green!50}$0.58{\pm \scriptstyle 0.02}$ & \cellcolor{Green!44}$0.52{\pm \scriptstyle 0.18}$ & \cellcolor{Green!23}$0.27{\pm \scriptstyle 0.08}$ & \cellcolor{Green!0}$0.01{\pm \scriptstyle 0.01}$ & \cellcolor{Green!4}$0.06{\pm \scriptstyle 0.03}$ \\
jnk3 & \cellcolor{Green!28}$0.10{\pm \scriptstyle 0.03}$ & \cellcolor{Green!29}$0.10{\pm \scriptstyle 0.02}$ & \cellcolor{Green!4}$0.06{\pm \scriptstyle 0.01}$ & \cellcolor{Green!29}$0.10{\pm \scriptstyle 0.03}$ & \cellcolor{Green!20}$0.08{\pm \scriptstyle 0.02}$ & \cellcolor{Green!20}$0.08{\pm \scriptstyle 0.04}$ & \cellcolor{Green!6}$0.06{\pm \scriptstyle 0.02}$ & \cellcolor{Green!0}$0.05{\pm \scriptstyle 0.04}$ & \cellcolor{Green!6}$0.06{\pm \scriptstyle 0.01}$ \\
median\_1 & \cellcolor{Green!36}$0.13{\pm \scriptstyle 0.03}$ & \cellcolor{Green!33}$0.13{\pm \scriptstyle 0.00}$ & \cellcolor{Green!48}$0.17{\pm \scriptstyle 0.02}$ & \cellcolor{Green!36}$0.13{\pm \scriptstyle 0.02}$ & \cellcolor{Green!44}$0.16{\pm \scriptstyle 0.01}$ & \cellcolor{Green!42}$0.15{\pm \scriptstyle 0.04}$ & \cellcolor{Green!50}$0.17{\pm \scriptstyle 0.03}$ & \cellcolor{Green!1}$0.03{\pm \scriptstyle 0.01}$ & \cellcolor{Green!0}$0.02{\pm \scriptstyle 0.00}$ \\
median\_2 & \cellcolor{Green!31}$0.08{\pm \scriptstyle 0.01}$ & \cellcolor{Green!29}$0.08{\pm \scriptstyle 0.02}$ & \cellcolor{Green!42}$0.11{\pm \scriptstyle 0.01}$ & \cellcolor{Green!47}$0.12{\pm \scriptstyle 0.01}$ & \cellcolor{Green!50}$0.12{\pm \scriptstyle 0.02}$ & \cellcolor{Green!48}$0.12{\pm \scriptstyle 0.01}$ & \cellcolor{Green!38}$0.10{\pm \scriptstyle 0.01}$ & \cellcolor{Green!1}$0.01{\pm \scriptstyle 0.00}$ & \cellcolor{Green!0}$0.01{\pm \scriptstyle 0.00}$ \\
mestranol\_similarity & \cellcolor{Green!50}$0.37{\pm \scriptstyle 0.01}$ & \cellcolor{Green!37}$0.28{\pm \scriptstyle 0.03}$ & \cellcolor{Green!49}$0.36{\pm \scriptstyle 0.05}$ & \cellcolor{Green!46}$0.34{\pm \scriptstyle 0.00}$ & \cellcolor{Green!47}$0.35{\pm \scriptstyle 0.03}$ & \cellcolor{Green!48}$0.36{\pm \scriptstyle 0.03}$ & \cellcolor{Green!43}$0.32{\pm \scriptstyle 0.09}$ & \cellcolor{Green!0}$0.01{\pm \scriptstyle 0.00}$ & \cellcolor{Green!0}$0.02{\pm \scriptstyle 0.00}$ \\
osimetrinib\_mpo & \cellcolor{Green!0}$0.00{\pm \scriptstyle 0.00}$ & \cellcolor{Green!46}$0.62{\pm \scriptstyle 0.01}$ & \cellcolor{Green!25}$0.34{\pm \scriptstyle 0.28}$ & \cellcolor{Green!48}$0.65{\pm \scriptstyle 0.03}$ & \cellcolor{Green!48}$0.65{\pm \scriptstyle 0.06}$ & \cellcolor{Green!47}$0.64{\pm \scriptstyle 0.00}$ & \cellcolor{Green!33}$0.45{\pm \scriptstyle 0.39}$ & \cellcolor{Green!22}$0.30{\pm \scriptstyle 0.31}$ & \cellcolor{Green!44}$0.59{\pm \scriptstyle 0.04}$ \\
perindopril\_mpo & \cellcolor{Green!25}$0.12{\pm \scriptstyle 0.01}$ & \cellcolor{Green!20}$0.10{\pm \scriptstyle 0.08}$ & \cellcolor{Green!38}$0.18{\pm \scriptstyle 0.04}$ & \cellcolor{Green!29}$0.14{\pm \scriptstyle 0.00}$ & \cellcolor{Green!29}$0.14{\pm \scriptstyle 0.01}$ & \cellcolor{Green!50}$0.23{\pm \scriptstyle 0.10}$ & \cellcolor{Green!43}$0.20{\pm \scriptstyle 0.08}$ & \cellcolor{Green!0}$0.00{\pm \scriptstyle 0.00}$ & \cellcolor{Green!0}$0.00{\pm \scriptstyle 0.00}$ \\
ranolazine\_mpo & \cellcolor{Green!41}$0.58{\pm \scriptstyle 0.09}$ & \cellcolor{Green!39}$0.54{\pm \scriptstyle 0.06}$ & \cellcolor{Green!41}$0.57{\pm \scriptstyle 0.09}$ & \cellcolor{Green!45}$0.62{\pm \scriptstyle 0.13}$ & \cellcolor{Green!50}$0.69{\pm \scriptstyle 0.08}$ & \cellcolor{Green!39}$0.55{\pm \scriptstyle 0.07}$ & \cellcolor{Green!20}$0.29{\pm \scriptstyle 0.05}$ & \cellcolor{Green!0}$0.00{\pm \scriptstyle 0.00}$ & \cellcolor{Green!8}$0.11{\pm \scriptstyle 0.02}$ \\
rdkit\_logp & \cellcolor{Green!50}$20.85{\pm \scriptstyle 0.21}$ & \cellcolor{Green!32}$14.76{\pm \scriptstyle 0.45}$ & \cellcolor{Green!49}$20.67{\pm \scriptstyle 0.66}$ & \cellcolor{Green!49}$20.73{\pm \scriptstyle 1.05}$ & \cellcolor{Green!44}$18.74{\pm \scriptstyle 1.83}$ & \cellcolor{Green!38}$16.94{\pm \scriptstyle 7.17}$ & \cellcolor{Green!48}$20.15{\pm \scriptstyle 3.07}$ & \cellcolor{Green!0}$3.12{\pm \scriptstyle 1.20}$ & \cellcolor{Green!6}$5.49{\pm \scriptstyle 3.01}$ \\
rdkit\_qed & \cellcolor{Green!50}$0.82{\pm \scriptstyle 0.00}$ & \cellcolor{Green!7}$0.56{\pm \scriptstyle 0.02}$ & \cellcolor{Green!49}$0.82{\pm \scriptstyle 0.11}$ & \cellcolor{Green!42}$0.77{\pm \scriptstyle 0.02}$ & \cellcolor{Green!41}$0.77{\pm \scriptstyle 0.10}$ & \cellcolor{Green!46}$0.80{\pm \scriptstyle 0.10}$ & \cellcolor{Green!18}$0.63{\pm \scriptstyle 0.03}$ & \cellcolor{Green!0}$0.52{\pm \scriptstyle 0.09}$ & \cellcolor{Green!12}$0.60{\pm \scriptstyle 0.05}$ \\
sa\_tdc & \cellcolor{Green!20}$7.64{\pm \scriptstyle 0.47}$ & \cellcolor{Green!50}$8.74{\pm \scriptstyle 0.11}$ & \cellcolor{Green!3}$7.00{\pm \scriptstyle 0.68}$ & \cellcolor{Green!29}$7.98{\pm \scriptstyle 0.02}$ & \cellcolor{Green!0}$6.86{\pm \scriptstyle 1.96}$ & \cellcolor{Green!29}$7.96{\pm \scriptstyle 0.05}$ & \cellcolor{Green!8}$7.16{\pm \scriptstyle 0.30}$ & \cellcolor{Green!39}$8.36{\pm \scriptstyle 0.46}$ & \cellcolor{Green!45}$8.59{\pm \scriptstyle 0.13}$ \\
scaffold\_hop & \cellcolor{Green!4}$0.34{\pm \scriptstyle 0.00}$ & \cellcolor{Green!18}$0.36{\pm \scriptstyle 0.01}$ & \cellcolor{Green!37}$0.37{\pm \scriptstyle 0.00}$ & \cellcolor{Green!33}$0.37{\pm \scriptstyle 0.00}$ & \cellcolor{Green!32}$0.37{\pm \scriptstyle 0.00}$ & \cellcolor{Green!36}$0.37{\pm \scriptstyle 0.01}$ & \cellcolor{Green!50}$0.38{\pm \scriptstyle 0.01}$ & \cellcolor{Green!0}$0.34{\pm \scriptstyle 0.01}$ & \cellcolor{Green!4}$0.34{\pm \scriptstyle 0.00}$ \\
sitagliptin\_mpo & \cellcolor{Green!16}$0.05{\pm \scriptstyle 0.03}$ & \cellcolor{Green!5}$0.02{\pm \scriptstyle 0.01}$ & \cellcolor{Green!25}$0.08{\pm \scriptstyle 0.07}$ & \cellcolor{Green!25}$0.08{\pm \scriptstyle 0.06}$ & \cellcolor{Green!26}$0.08{\pm \scriptstyle 0.06}$ & \cellcolor{Green!50}$0.15{\pm \scriptstyle 0.13}$ & \cellcolor{Green!1}$0.00{\pm \scriptstyle 0.01}$ & \cellcolor{Green!0}$0.00{\pm \scriptstyle 0.00}$ & \cellcolor{Green!0}$0.00{\pm \scriptstyle 0.00}$ \\
thiothixene\_rediscovery & \cellcolor{Green!33}$0.15{\pm \scriptstyle 0.02}$ & \cellcolor{Green!28}$0.13{\pm \scriptstyle 0.03}$ & \cellcolor{Green!41}$0.18{\pm \scriptstyle 0.02}$ & \cellcolor{Green!50}$0.21{\pm \scriptstyle 0.04}$ & \cellcolor{Green!43}$0.18{\pm \scriptstyle 0.05}$ & \cellcolor{Green!43}$0.18{\pm \scriptstyle 0.03}$ & \cellcolor{Green!33}$0.15{\pm \scriptstyle 0.01}$ & \cellcolor{Green!0}$0.02{\pm \scriptstyle 0.01}$ & \cellcolor{Green!2}$0.03{\pm \scriptstyle 0.01}$ \\
troglitazone\_rediscovery & \cellcolor{Green!28}$0.10{\pm \scriptstyle 0.01}$ & \cellcolor{Green!30}$0.11{\pm \scriptstyle 0.02}$ & \cellcolor{Green!50}$0.17{\pm \scriptstyle 0.02}$ & \cellcolor{Green!44}$0.15{\pm \scriptstyle 0.00}$ & \cellcolor{Green!46}$0.16{\pm \scriptstyle 0.01}$ & \cellcolor{Green!42}$0.14{\pm \scriptstyle 0.01}$ & \cellcolor{Green!34}$0.12{\pm \scriptstyle 0.02}$ & \cellcolor{Green!0}$0.02{\pm \scriptstyle 0.01}$ & \cellcolor{Green!0}$0.02{\pm \scriptstyle 0.00}$ \\
valsartan\_smarts & \cellcolor{Green!0}$0.00{\pm \scriptstyle 0.00}$ & \cellcolor{Green!0}$0.00{\pm \scriptstyle 0.00}$ & \cellcolor{Green!0}$0.00{\pm \scriptstyle 0.00}$ & \cellcolor{Green!0}$0.00{\pm \scriptstyle 0.00}$ & \cellcolor{Green!0}$0.00{\pm \scriptstyle 0.00}$ & \cellcolor{Green!0}$0.00{\pm \scriptstyle 0.00}$ & \cellcolor{Green!0}$0.00{\pm \scriptstyle 0.00}$ & \cellcolor{Green!0}$0.00{\pm \scriptstyle 0.00}$ & \cellcolor{Green!0}$0.00{\pm \scriptstyle 0.00}$ \\
zaleplon\_mpo & \cellcolor{Green!4}$0.02{\pm \scriptstyle 0.02}$ & \cellcolor{Green!5}$0.02{\pm \scriptstyle 0.01}$ & \cellcolor{Green!28}$0.12{\pm \scriptstyle 0.05}$ & \cellcolor{Green!34}$0.15{\pm \scriptstyle 0.10}$ & \cellcolor{Green!50}$0.22{\pm \scriptstyle 0.11}$ & \cellcolor{Green!26}$0.12{\pm \scriptstyle 0.04}$ & \cellcolor{Green!12}$0.05{\pm \scriptstyle 0.04}$ & \cellcolor{Green!0}$0.00{\pm \scriptstyle 0.00}$ & \cellcolor{Green!0}$0.00{\pm \scriptstyle 0.00}$ \\
\midrule
Sum (normalized per row) & \cellcolor{Green!29}$12.35{\pm \scriptstyle 1.33}$ & \cellcolor{Green!33}$13.74{\pm \scriptstyle 1.26}$ & \cellcolor{Green!43}$17.28{\pm \scriptstyle 2.56}$ & \cellcolor{Green!49}$19.54{\pm \scriptstyle 1.81}$ & \cellcolor{Green!50}$19.54{\pm \scriptstyle 4.59}$ & \cellcolor{Green!49}$19.24{\pm \scriptstyle 8.38}$ & \cellcolor{Green!35}$14.52{\pm \scriptstyle 4.84}$ & \cellcolor{Green!0}$1.65{\pm \scriptstyle 2.47}$ & \cellcolor{Green!6}$3.87{\pm \scriptstyle 3.57}$ \\
\bottomrule
\end{tabular}
}
    \caption{Results on the PMO benchmark for a 2-dimensional latent space. The best output of the optimization campaign over max. 100 iterations are averaged over three runs, using 10 initial SOBOL-sampled points. The last row is computed as in \ref{tab:results:absolute_values_for_128_latent_dim}. \texttt{Bounce} and \texttt{ProbRep}'s underlying results are exactly the same as in the 128D case, but restricted to 100 iterations. We note that \texttt{BAxUS} defaults to \texttt{Turbo} in 2 dimensions.
    }
    \label{tab:results:absolute_values_for_2_dim_latent_space}
    \vspace{-0.75cm}
\end{table}

To benchmark the performance of HDBO on discrete sequences in continuous learned representations, we consider the PMO benchmark \citep{Gao:PMOMolOpt:2022,Huang:TDC:2021, Brown:Guacamol:2019}.\footnote{An introduction to PMO can be found in \cref{sec:appendix:PMO}} 
We select  Hvarfner's \texttt{VanillaBO}, \texttt{RandomLineBO}, \texttt{Turbo}, \texttt{BAxUS}, \texttt{SAASBO}, \texttt{Bounce}, and \texttt{ProbRep}, including also \texttt{HillClimbing}, \texttt{CMA-ES} and \texttt{GeneticAlgorithm} as baselines. All solvers start from the same initial data to ensure a fair comparison except for \texttt{BAxUS} and \texttt{Bounce} due to the nature of their implementations.
%
We test the aforementioned methods on PMO \citep{Gao:PMOMolOpt:2022,Huang:TDC:2021}, which requires a discrete representation of small molecules. 
Thus, we train two MLP VAEs on SELFIES representations of small molecules using Zinc250k \citep{Irwin:ZINC20:2020, Zhu:TorchDrug:2022}. These generative models had 2 and 128 latent dimensions, allowing us to get an impression of how these models scale with dimensionality.
We restrict sequences to be of length 70 (adding \texttt{[nop]} tokens for padding); the post-processing renders an alphabet of 64 SELFIES tokens. Details can be found in Sec.~\ref{sec:appendix:training-vaes-on-selfies}.


The average best result over 3 runs (of maximum 300 iterations each) is presented in Tables~\ref{tab:results:absolute_values_for_128_latent_dim} and ~\ref{tab:results:absolute_values_for_2_dim_latent_space} for 128D and 2D latent spaces respectively. We see a clear advantage in the optimizers that work on learned representations, instead of in discrete space. Such a discrepancy is to be expected: methods that optimize in latent space have been presented with information prior to their optimization campaigns, while methods like \texttt{Bounce} and \texttt{ProbRep} explore the whole discrete space. Further, the simple baseline is reliably beaten by the continuous alternatives in lower dimensions except \texttt{Turbo}, but this advantage is not as clear in the 128D case, signaling a more complex problem. Some of these tasks, however, are equally challenging for all solvers. \texttt{deco\_hop} remains close to the original default value of 0.5, and there is no improvement over \texttt{valsartan\_smarts} (which only REINVENT improves on in the original PMO results \citep{Gao:PMOMolOpt:2022,Loeffler:REINVENT:2024}).



\section{Conclusion}

In this paper, we surveyed the field of high-dimensional Bayesian optimization (HDBO) focusing on discrete problems. This highlighted the need for (i) a novel taxonomy of the field that emphasizes the differences between methods that rely on unsupervised discrete information, and methods that optimize sequences directly, and (ii) a standardized framework for benchmarking HDBO methods. We approach these in the form of two software tools: \href{https://github.com/MachineLearningLifeScience/poli}{\texttt{poli}} and \href{https://github.com/MachineLearningLifeScience/poli-baselines}{\texttt{poli-baselines}}.
Using these tools, we implemented several HDBO methods and tested them in a standard benchmark for sequence design and small molecule optimization. Our findings suggest that optimizers that work on discrete sequence space (i.e. the \textit{structured spaces} family) perform as expected on problems of low complexity, but do not scale gracefully to problems with larger dictionaries/higher sequence lengths. In such cases, optimizers that leverage pre-trained latent-variable models have an edge over the other tested methods that work directly on sequence space. That being said, we find that, for low budgets, simpler baselines tend to perform as well or better than most BO methods, echoing recent criticisms of molecular optimization \citep{Tripp:MolGA:2023}.
Our framework opens the door to fair and easily-replicable comparisons. We expect \texttt{poli-baselines} to be used by practitioners for running HDBO solvers in up-to-date environments compared across several tasks in our ongoing benchmark, which we plan to expand to other discrete objectives in \texttt{poli}.
%

\paragraph{Limitations and societal impact.} Although we taxonomize different families of HDBO methods, we have only benchmarked a subset; moreover, the black boxes we use are often simple, closed-form functions or data-driven oracles, limiting the applicability of our benchmark to real-world scenarios. The field of discrete sequence optimization for biology/chemistry needs better black box oracles with more relevance to their respective domains. Another limitation of our approach is the lack of a systematic analysis on the myriad of design choices made on these optimization campaigns; we strived for documenting them properly, but an in-depth analysis is called for in future work. Our benchmark is ongoing and we plan to include further experiments in the project's website with, hopefully, participation from the community.
Finally, we note that optimizing small molecules or proteins opens the door to both drug discovery, but also dual use \citep{Urbina:DualUse:2022}.

\clearpage
\begin{ack}
The work was partly funded by the Novo Nordisk Foundation through the Center for Basic Machine Learning Research in Life Science (NNF20OC0062606). RM is funded by the Danish Data Science Academy, which is funded by the Novo Nordisk Foundation (NNF21SA0069429) and VILLUM FONDEN (40516). SH was further supported by a research grant (42062) from VILLUM FONDEN as well as funding from the European Research Council (ERC) under the European Union’s Horizon programme (grant agreement 101125993). WB was supported by VILLUM FONDEN (40578).
This work was in part supported by the Pioneer Centre for AI (DRNF grant number P1). MGD thanks Sergio Garrido and Anshuk Uppal for feedback on early versions of this document; Peter Mørch Groth for useful discussions; Luigi Nardi, Erik Hellsten and Raffaello Baluyot for feedback on the preprint version of this paper, and Samuel Stanton for feedback on the software. 
\end{ack}

\bibliographystyle{apalike} 
\bibliography{references} 

\begin{thebibliography}{}

\bibitem[Ahmed et~al., 2016]{Ahmed:FOBO:2016}
Ahmed, M.~O., Shahriari, B., and Schmidt, M. (2016).
\newblock Do we need “harmless” bayesian optimization and “first-order” bayesian optimization.
\newblock {\em NIPS BayesOpt}.

\bibitem[Ament and Gomes, 2022]{Ament:SBOSDA:2022}
Ament, S.~E. and Gomes, C.~P. (2022).
\newblock Scalable first-order {B}ayesian optimization via structured automatic differentiation.
\newblock In Chaudhuri, K., Jegelka, S., Song, L., Szepesvari, C., Niu, G., and Sabato, S., editors, {\em Proceedings of the 39th International Conference on Machine Learning}, volume 162 of {\em Proceedings of Machine Learning Research}, pages 500--516. PMLR.

\bibitem[Antonov et~al., 2022]{Antonov:KPCABO2022}
Antonov, K., Raponi, E., Wang, H., and Doerr, C. (2022).
\newblock High dimensional bayesian optimization with kernel principal component analysis.
\newblock In {\em International Conference on Parallel Problem Solving from Nature}, pages 118--131. Springer.

\bibitem[Balandat et~al., 2020]{Balandat:BoTorch:2020}
Balandat, M., Karrer, B., Jiang, D., Daulton, S., Letham, B., Wilson, A.~G., and Bakshy, E. (2020).
\newblock Botorch: A framework for efficient monte-carlo bayesian optimization.
\newblock In Larochelle, H., Ranzato, M., Hadsell, R., Balcan, M., and Lin, H., editors, {\em Advances in Neural Information Processing Systems}, volume~33, pages 21524--21538. Curran Associates, Inc.

\bibitem[Baptista and Poloczek, 2018]{Baptista:Combinatorial:2018}
Baptista, R. and Poloczek, M. (2018).
\newblock {B}ayesian optimization of combinatorial structures.
\newblock In Dy, J. and Krause, A., editors, {\em Proceedings of the 35th International Conference on Machine Learning}, volume~80 of {\em Proceedings of Machine Learning Research}, pages 462--471. PMLR.

\bibitem[Binois et~al., 2015]{Binois:WarpedREMBO:2015}
Binois, M., Ginsbourger, D., and Roustant, O. (2015).
\newblock A warped kernel improving robustness in bayesian optimization via random embeddings.
\newblock In {\em Learning and Intelligent Optimization: 9th International Conference, LION 9, Lille, France, January 12-15, 2015. Revised Selected Papers 9}, pages 281--286. Springer.

\bibitem[Binois et~al., 2020]{Binois:REMBO2:2020}
Binois, M., Ginsbourger, D., and Roustant, O. (2020).
\newblock On the choice of the low-dimensional domain for global optimization via random embeddings.
\newblock {\em Journal of global optimization}, 76(1):69--90.

\bibitem[Binois and Wycoff, 2022]{BinoisWycoff:HDGPs:2022}
Binois, M. and Wycoff, N. (2022).
\newblock A survey on high-dimensional gaussian process modeling with application to bayesian optimization.
\newblock {\em ACM Transactions on Evolutionary Learning and Optimization}, 2(2).

\bibitem[Blaabjerg et~al., 2023]{Blaabjerg:RASP:2023}
Blaabjerg, L.~M., Kassem, M.~M., Good, L.~L., Jonsson, N., Cagiada, M., Johansson, K.~E., Boomsma, W., Stein, A., and Lindorff-Larsen, K. (2023).
\newblock Rapid protein stability prediction using deep learning representations.
\newblock {\em eLife}, 12:e82593.

\bibitem[Borovitskiy et~al., 2020]{Borovitsky:MaternKernelOnManifold:2020}
Borovitskiy, V., Terenin, A., Mostowsky, P., and Deisenroth~(he/him), M. (2020).
\newblock Mat\'{e}rn gaussian processes on riemannian manifolds.
\newblock In Larochelle, H., Ranzato, M., Hadsell, R., Balcan, M., and Lin, H., editors, {\em Advances in Neural Information Processing Systems}, volume~33, pages 12426--12437. Curran Associates, Inc.

\bibitem[Boumal, 2023]{Boumal:ManifoldOpt:2023}
Boumal, N. (2023).
\newblock {\em An introduction to optimization on smooth manifolds}.
\newblock Cambridge University Press.

\bibitem[Bridges et~al., 2019]{Bridges:ActiveManifolds:2019}
Bridges, R.~A., Gruber, A.~D., Felder, C., Verma, M., and Hoff, C. (2019).
\newblock Active manifolds: A non-linear analogue to active subspaces.

\bibitem[Brown et~al., 2019]{Brown:Guacamol:2019}
Brown, N., Fiscato, M., Segler, M.~H., and Vaucher, A.~C. (2019).
\newblock Guacamol: benchmarking models for de novo molecular design.
\newblock {\em Journal of chemical information and modeling}, 59(3):1096--1108.

\bibitem[Chapman and Chang, 2000]{Chapman:Biopython:2000}
Chapman, B. and Chang, J. (2000).
\newblock Biopython: Python tools for computational biology.
\newblock {\em SIGBIO Newsl.}, 20(2):15–19.

\bibitem[Chen et~al., 2012]{Chen:HDS:2012}
Chen, B., Castro, R.~M., and Krause, A. (2012).
\newblock Joint optimization and variable selection of high-dimensional gaussian processes.
\newblock In {\em Proceedings of the 29th International Conference on Machine Learning (ICML)}.

\bibitem[Chen et~al., 2020]{Jingfan:SILBOHDBO:2020}
Chen, J., Zhu, G., Yuan, C., and Huang, Y. (2020).
\newblock Semi-supervised embedding learning for high-dimensional bayesian optimization.

\bibitem[Chen et~al., 2024]{Chen:PGLBO:2024}
Chen, T., Duan, Y., Li, D., Qi, L., Shi, Y., and Gao, Y. (2024).
\newblock Pg-lbo: Enhancing high-dimensional bayesian optimization with pseudo-label and gaussian process guidance.
\newblock In {\em Proceedings of the AAAI Conference on Artificial Intelligence}, volume~38, pages 11381--11389.

\bibitem[Cowen-Rivers et~al., 2022]{Cowen:HEBO:2022}
Cowen-Rivers, A.~I., Lyu, W., Tutunov, R., Wang, Z., Grosnit, A., Griffiths, R.~R., Maraval, A.~M., Jianye, H., Wang, J., Peters, J., et~al. (2022).
\newblock Hebo: Pushing the limits of sample-efficient hyper-parameter optimisation.
\newblock {\em Journal of Artificial Intelligence Research}, 74:1269--1349.

\bibitem[Daulton et~al., 2022a]{Daulton:MORBO:2022}
Daulton, S., Eriksson, D., Balandat, M., and Bakshy, E. (2022a).
\newblock Multi-objective bayesian optimization over high-dimensional search spaces.
\newblock In Cussens, J. and Zhang, K., editors, {\em Proceedings of the Thirty-Eighth Conference on Uncertainty in Artificial Intelligence}, volume 180 of {\em Proceedings of Machine Learning Research}, pages 507--517. PMLR.

\bibitem[Daulton et~al., 2022b]{Daulton:PR:2022}
Daulton, S., Wan, X., Eriksson, D., Balandat, M., Osborne, M.~A., and Bakshy, E. (2022b).
\newblock Bayesian optimization over discrete and mixed spaces via probabilistic reparameterization.
\newblock In Koyejo, S., Mohamed, S., Agarwal, A., Belgrave, D., Cho, K., and Oh, A., editors, {\em Advances in Neural Information Processing Systems}, volume~35, pages 12760--12774. Curran Associates, Inc.

\bibitem[Delbridge et~al., 2020]{Delbridge:RPAGP:2020}
Delbridge, I., Bindel, D., and Wilson, A.~G. (2020).
\newblock Randomly projected additive {G}aussian processes for regression.
\newblock In III, H.~D. and Singh, A., editors, {\em Proceedings of the 37th International Conference on Machine Learning}, volume 119 of {\em Proceedings of Machine Learning Research}, pages 2453--2463. PMLR.

\bibitem[Delgado et~al., 2019]{Delgado:FOLDX5:2019}
Delgado, J., Radusky, L.~G., Cianferoni, D., and Serrano, L. (2019).
\newblock {FoldX 5.0: working with RNA, small molecules and a new graphical interface}.
\newblock {\em Bioinformatics}, 35(20):4168--4169.

\bibitem[Deshwal et~al., 2023]{Deshwal:BODI:2023}
Deshwal, A., Ament, S., Balandat, M., Bakshy, E., Doppa, J.~R., and Eriksson, D. (2023).
\newblock Bayesian optimization over high-dimensional combinatorial spaces via dictionary-based embeddings.
\newblock In Ruiz, F., Dy, J., and van~de Meent, J.-W., editors, {\em Proceedings of The 26th International Conference on Artificial Intelligence and Statistics}, volume 206 of {\em Proceedings of Machine Learning Research}, pages 7021--7039. PMLR.

\bibitem[Deshwal et~al., 2021a]{Deshwal:HybridSpaces:2021}
Deshwal, A., Belakaria, S., and Doppa, J.~R. (2021a).
\newblock Bayesian optimization over hybrid spaces.
\newblock In Meila, M. and Zhang, T., editors, {\em Proceedings of the 38th International Conference on Machine Learning}, volume 139 of {\em Proceedings of Machine Learning Research}, pages 2632--2643. PMLR.

\bibitem[Deshwal et~al., 2021b]{Deshwal:MercBO:2021}
Deshwal, A., Belakaria, S., and Doppa, J.~R. (2021b).
\newblock Mercer features for efficient combinatorial bayesian optimization.
\newblock In {\em Proceedings of the AAAI Conference on Artificial Intelligence}, volume~35, pages 7210--7218.

\bibitem[Deshwal and Doppa, 2021]{Deshwal:LADDER:2021}
Deshwal, A. and Doppa, J. (2021).
\newblock Combining latent space and structured kernels for bayesian optimization over combinatorial spaces.
\newblock In Ranzato, M., Beygelzimer, A., Dauphin, Y., Liang, P., and Vaughan, J.~W., editors, {\em Advances in Neural Information Processing Systems}, volume~34, pages 8185--8200. Curran Associates, Inc.

\bibitem[Diouane et~al., 2023]{Diouane:TREGO:2023}
Diouane, Y., Picheny, V., Riche, R.~L., and Perrotolo, A. S.~D. (2023).
\newblock Trego: a trust-region framework for efficient global optimization.
\newblock {\em Journal of Global Optimization}, 86(1):1--23.

\bibitem[Djolonga et~al., 2013]{Djolonga:HDBandits:2013}
Djolonga, J., Krause, A., and Cevher, V. (2013).
\newblock High-dimensional gaussian process bandits.
\newblock In Burges, C., Bottou, L., Welling, M., Ghahramani, Z., and Weinberger, K., editors, {\em Advances in Neural Information Processing Systems}, volume~26. Curran Associates, Inc.

\bibitem[Dreczkowski et~al., 2023]{dreczkowski:mcbo:2023}
Dreczkowski, K., Grosnit, A., and Ammar, H.~B. (2023).
\newblock Framework and benchmarks for combinatorial and mixed-variable bayesian optimization.

\bibitem[Eissman et~al., 2018]{Eissman:AttrAdjust:2018}
Eissman, S., Levy, D., Shu, R., Bartzsch, S., and Ermon, S. (2018).
\newblock Bayesian optimization and attribute adjustment.
\newblock In {\em Proc. 34th Conference on Uncertainty in Artificial Intelligence}.

\bibitem[Eriksson et~al., 2018]{Eriksson:DSKIP:2018}
Eriksson, D., Dong, K., Lee, E., Bindel, D., and Wilson, A.~G. (2018).
\newblock Scaling gaussian process regression with derivatives.
\newblock In Bengio, S., Wallach, H., Larochelle, H., Grauman, K., Cesa-Bianchi, N., and Garnett, R., editors, {\em Advances in Neural Information Processing Systems}, volume~31. Curran Associates, Inc.

\bibitem[Eriksson and Jankowiak, 2021]{Eriksson:SAASBO:2021}
Eriksson, D. and Jankowiak, M. (2021).
\newblock High-dimensional {Bayesian} optimization with sparse axis-aligned subspaces.
\newblock In de~Campos, C. and Maathuis, M.~H., editors, {\em Proceedings of the Thirty-Seventh Conference on Uncertainty in Artificial Intelligence}, volume 161 of {\em Proceedings of Machine Learning Research}, pages 493--503. PMLR.

\bibitem[Eriksson et~al., 2019]{Eriksson:TuRBO:2019}
Eriksson, D., Pearce, M., Gardner, J., Turner, R.~D., and Poloczek, M. (2019).
\newblock Scalable global optimization via local bayesian optimization.
\newblock In Wallach, H., Larochelle, H., Beygelzimer, A., d\textquotesingle Alch\'{e}-Buc, F., Fox, E., and Garnett, R., editors, {\em Advances in Neural Information Processing Systems}, volume~32. Curran Associates, Inc.

\bibitem[Feragen et~al., 2015]{Feragen:GeodesicKernels:2015}
Feragen, A., Lauze, F., and Hauberg, S. (2015).
\newblock Geodesic {Exponential} {Kernels}: {When} {Curvature} and {Linearity} {Conflict}.
\newblock In {\em Proceedings of the IEEE conference on computer vision and pattern recognition}, pages 3032--3042.

\bibitem[Gao et~al., 2022]{Gao:PMOMolOpt:2022}
Gao, W., Fu, T., Sun, J., and Coley, C. (2022).
\newblock Sample efficiency matters: A benchmark for practical molecular optimization.
\newblock In Koyejo, S., Mohamed, S., Agarwal, A., Belgrave, D., Cho, K., and Oh, A., editors, {\em Advances in Neural Information Processing Systems}, volume~35, pages 21342--21357. Curran Associates, Inc.

\bibitem[Garc{\'\i}a-Orteg{\'o}n et~al., 2022]{Garcia:DOCKSTRING:2022}
Garc{\'\i}a-Orteg{\'o}n, M., Simm, G.~N., Tripp, A.~J., Hern{\'a}ndez-Lobato, J.~M., Bender, A., and Bacallado, S. (2022).
\newblock Dockstring: easy molecular docking yields better benchmarks for ligand design.
\newblock {\em Journal of chemical information and modeling}, 62(15):3486--3502.

\bibitem[Gardner et~al., 2017]{Gardner:AddStruct:2017}
Gardner, J., Guo, C., Weinberger, K., Garnett, R., and Grosse, R. (2017).
\newblock {Discovering and Exploiting Additive Structure for Bayesian Optimization}.
\newblock In Singh, A. and Zhu, J., editors, {\em Proceedings of the 20th International Conference on Artificial Intelligence and Statistics}, volume~54 of {\em Proceedings of Machine Learning Research}, pages 1311--1319. PMLR.

\bibitem[Gardner et~al., 2018]{Gardner:Gpytorch:2018}
Gardner, J., Pleiss, G., Weinberger, K.~Q., Bindel, D., and Wilson, A.~G. (2018).
\newblock Gpytorch: Blackbox matrix-matrix gaussian process inference with gpu acceleration.
\newblock In Bengio, S., Wallach, H., Larochelle, H., Grauman, K., Cesa-Bianchi, N., and Garnett, R., editors, {\em Advances in Neural Information Processing Systems}, volume~31. Curran Associates, Inc.

\bibitem[Garnett, 2023]{Garnett2023BObook}
Garnett, R. (2023).
\newblock {\em {Bayesian Optimization}}.
\newblock Cambridge University Press.

\bibitem[Garnett et~al., 2013]{Garnett:ActiveLLEmb:2013}
Garnett, R., Osborne, M.~A., and Hennig, P. (2013).
\newblock Active learning of linear embeddings for gaussian processes.

\bibitem[Garrido-Merchán and Hernández-Lobato, 2020]{GarridoMerchan:CatIntBO:2020}
Garrido-Merchán, E.~C. and Hernández-Lobato, D. (2020).
\newblock Dealing with categorical and integer-valued variables in bayesian optimization with gaussian processes.
\newblock {\em Neurocomputing}, 380:20--35.

\bibitem[Goodfellow et~al., 2014]{Goodfellow:GAN:2014}
Goodfellow, I., Pouget-Abadie, J., Mirza, M., Xu, B., Warde-Farley, D., Ozair, S., Courville, A., and Bengio, Y. (2014).
\newblock Generative adversarial nets.
\newblock In Ghahramani, Z., Welling, M., Cortes, C., Lawrence, N., and Weinberger, K., editors, {\em Advances in Neural Information Processing Systems}, volume~27.

\bibitem[Griffiths and Hernández-Lobato, 2020]{Griffiths:ConstrainedBOVAEs:2020}
Griffiths, R.-R. and Hernández-Lobato, J.~M. (2020).
\newblock Constrained bayesian optimization for automatic chemical design using variational autoencoders.
\newblock {\em Chemical Science}, 11(2):577–586.

\bibitem[Griffiths et~al., 2023]{Griffiths:GAUCHE:2023}
Griffiths, R.-R., Klarner, L., Moss, H., Ravuri, A., Truong, S., Du, Y., Stanton, S., Tom, G., Rankovic, B., Jamasb, A., Deshwal, A., Schwartz, J., Tripp, A., Kell, G., Frieder, S., Bourached, A., Chan, A., Moss, J., Guo, C., D\"{u}rholt, J.~P., Chaurasia, S., Park, J.~W., Strieth-Kalthoff, F., Lee, A., Cheng, B., Aspuru-Guzik, A., Schwaller, P., and Tang, J. (2023).
\newblock Gauche: A library for gaussian processes in chemistry.
\newblock In Oh, A., Naumann, T., Globerson, A., Saenko, K., Hardt, M., and Levine, S., editors, {\em Advances in Neural Information Processing Systems}, volume~36, pages 76923--76946. Curran Associates, Inc.

\bibitem[Grosnit et~al., 2021]{Grosnit:VAEDML:2021}
Grosnit, A., Tutunov, R., Maraval, A.~M., Griffiths, R.-R., Cowen-Rivers, A.~I., Yang, L., Zhu, L., Lyu, W., Chen, Z., Wang, J., Peters, J., and Bou-Ammar, H. (2021).
\newblock High-dimensional bayesian optimisation with variational autoencoders and deep metric learning.

\bibitem[Gruver et~al., 2023]{Gruver:LAMBO2:2023}
Gruver, N., Stanton, S., Frey, N., Rudner, T. G.~J., Hotzel, I., Lafrance-Vanasse, J., Rajpal, A., Cho, K., and Wilson, A.~G. (2023).
\newblock Protein design with guided discrete diffusion.
\newblock In Oh, A., Naumann, T., Globerson, A., Saenko, K., Hardt, M., and Levine, S., editors, {\em Advances in Neural Information Processing Systems}, volume~36, pages 12489--12517. Curran Associates, Inc.

\bibitem[Gómez-Bombarelli et~al., 2018]{Bombarelli:AutoChemDesign:2018}
Gómez-Bombarelli, R., Wei, J.~N., Duvenaud, D., Hernández-Lobato, J.~M., Sánchez-Lengeling, B., Sheberla, D., Aguilera-Iparraguirre, J., Hirzel, T.~D., Adams, R.~P., and Aspuru-Guzik, A. (2018).
\newblock Automatic chemical design using a data-driven continuous representation of molecules.
\newblock {\em ACS Central Science}, 4(2):268--276.
\newblock PMID: 29532027.

\bibitem[Han et~al., 2021]{Han:AddGPUCB:2021}
Han, E., Arora, I., and Scarlett, J. (2021).
\newblock High-dimensional bayesian optimization via tree-structured additive models.
\newblock {\em Proceedings of the AAAI Conference on Artificial Intelligence}, 35(9):7630--7638.

\bibitem[Harris et~al., 2020]{harris202numpy}
Harris, C.~R., Millman, K.~J., van~der Walt, S.~J., Gommers, R., Virtanen, P., Cournapeau, D., Wieser, E., Taylor, J., Berg, S., Smith, N.~J., Kern, R., Picus, M., Hoyer, S., van Kerkwijk, M.~H., Brett, M., Haldane, A., del R{\'{i}}o, J.~F., Wiebe, M., Peterson, P., G{\'{e}}rard-Marchant, P., Sheppard, K., Reddy, T., Weckesser, W., Abbasi, H., Gohlke, C., and Oliphant, T.~E. (2020).
\newblock Array programming with {NumPy}.
\newblock {\em Nature}, 585(7825):357--362.

\bibitem[Hebbal et~al., 2019]{Hebbal:DeepGPMO:2019}
Hebbal, A., Brevault, L., Balesdent, M., Talbi, E.-G., and Melab, N. (2019).
\newblock Multi-objective optimization using deep gaussian processes: application to aerospace vehicle design.
\newblock In {\em AIAA Scitech 2019 Forum}, page 1973.

\bibitem[Hebbal et~al., 2021]{Hebbal:DeepGPs:2021}
Hebbal, A., Brevault, L., Balesdent, M., Talbi, E.-G., and Melab, N. (2021).
\newblock Bayesian optimization using deep gaussian processes with applications to aerospace system design.
\newblock {\em Optimization and Engineering}, 22:321--361.

\bibitem[Hellsten et~al., 2023]{Hellsten:GTBO:2023}
Hellsten, E.~O., Hvarfner, C., Papenmeier, L., and Nardi, L. (2023).
\newblock High-dimensional bayesian optimization with group testing.

\bibitem[Horiguchi et~al., 2022]{Horiguchi:MahalaBatchBO:2022}
Horiguchi, S.~A., Iwata, T., Tsuzuki, T., and Ozawa, Y. (2022).
\newblock Linear embedding-based high-dimensional batch bayesian optimization without reconstruction mappings.

\bibitem[Hu et~al., 2024]{Hu:MAVEBO:2024}
Hu, S., Li, J., and Cai, Z. (2024).
\newblock An adaptive dimension reduction estimation method for high-dimensional bayesian optimization.

\bibitem[Huang et~al., 2021]{Huang:TDC:2021}
Huang, K., Fu, T., Gao, W., Zhao, Y., Roohani, Y.~H., Leskovec, J., Coley, C.~W., Xiao, C., Sun, J., and Zitnik, M. (2021).
\newblock Therapeutics data commons: Machine learning datasets and tasks for drug discovery and development.
\newblock In {\em Thirty-fifth Conference on Neural Information Processing Systems Datasets and Benchmarks Track (Round 1)}.

\bibitem[Hvarfner et~al., 2023]{Hvarfner:SCOREBO:2023}
Hvarfner, C., Hellsten, E., Hutter, F., and Nardi, L. (2023).
\newblock Self-correcting bayesian optimization through bayesian active learning.
\newblock In Oh, A., Naumann, T., Globerson, A., Saenko, K., Hardt, M., and Levine, S., editors, {\em Advances in Neural Information Processing Systems}, volume~36, pages 79173--79199. Curran Associates, Inc.

\bibitem[Hvarfner et~al., 2024]{Hvarfner:VanilaBO:2024}
Hvarfner, C., Hellsten, E.~O., and Nardi, L. (2024).
\newblock Vanilla bayesian optimization performs great in high dimensions.

\bibitem[Irwin et~al., 2020]{Irwin:ZINC20:2020}
Irwin, J.~J., Tang, K.~G., Young, J., Dandarchuluun, C., Wong, B.~R., Khurelbaatar, M., Moroz, Y.~S., Mayfield, J., and Sayle, R.~A. (2020).
\newblock Zinc20—a free ultralarge-scale chemical database for ligand discovery.
\newblock {\em Journal of chemical information and modeling}, 60(12):6065--6073.

\bibitem[Jaquier and Rozo, 2020]{Jaquier:HDGABO:2020}
Jaquier, N. and Rozo, L. (2020).
\newblock High-dimensional bayesian optimization via nested riemannian manifolds.
\newblock In Larochelle, H., Ranzato, M., Hadsell, R., Balcan, M., and Lin, H., editors, {\em Advances in Neural Information Processing Systems}, volume~33, pages 20939--20951. Curran Associates, Inc.

\bibitem[Jaquier et~al., 2020]{Jaquier:GABO:2020}
Jaquier, N., Rozo, L., Calinon, S., and B\"urger, M. (2020).
\newblock Bayesian optimization meets riemannian manifolds in robot learning.
\newblock In Kaelbling, L.~P., Kragic, D., and Sugiura, K., editors, {\em Proceedings of the Conference on Robot Learning}, volume 100 of {\em Proceedings of Machine Learning Research}, pages 233--246. PMLR.

\bibitem[Jin et~al., 2018]{Jin:JunctionTreeVAE:2018}
Jin, W., Barzilay, R., and Jaakkola, T. (2018).
\newblock Junction tree variational autoencoder for molecular graph generation.
\newblock In Dy, J. and Krause, A., editors, {\em Proceedings of the 35th International Conference on Machine Learning}, volume~80 of {\em Proceedings of Machine Learning Research}, pages 2323--2332.

\bibitem[Jones et~al., 1998]{Jones:BO:1998}
Jones, D.~R., Schonlau, M., and Welch, W.~J. (1998).
\newblock Efficient global optimization of expensive black-box functions.
\newblock {\em Journal of Global optimization}, 13:455--492.

\bibitem[Justesen et~al., 2020]{Justesen:DL4VGP:2020}
Justesen, N., Bontrager, P., Togelius, J., and Risi, S. (2020).
\newblock Deep learning for video game playing.
\newblock {\em IEEE Transactions on Games}, 12(1):1--20.

\bibitem[Khan et~al., 2023]{Khan:AntBO:2023}
Khan, A., Cowen-Rivers, A.~I., Grosnit, A., Robert, P.~A., Greiff, V., Smorodina, E., Rawat, P., Akbar, R., Dreczkowski, K., Tutunov, R., et~al. (2023).
\newblock Toward real-world automated antibody design with combinatorial bayesian optimization.
\newblock {\em Cell Reports Methods}, 3(1).

\bibitem[Kim et~al., 2022]{Kim:RMF:2022}
Kim, J., Choi, S., and Cho, M. (2022).
\newblock Combinatorial bayesian optimization with random mapping functions to convex polytopes.
\newblock In Cussens, J. and Zhang, K., editors, {\em Proceedings of the Thirty-Eighth Conference on Uncertainty in Artificial Intelligence}, volume 180 of {\em Proceedings of Machine Learning Research}, pages 1001--1011. PMLR.

\bibitem[Kingma and Welling, 2014]{Kingma:VAE:2014}
Kingma, D.~P. and Welling, M. (2014).
\newblock Auto-encoding variational bayes.
\newblock In Bengio, Y. and LeCun, Y., editors, {\em 2nd International Conference on Learning Representations, {ICLR} 2014, Banff, AB, Canada, April 14-16, 2014, Conference Track Proceedings}.

\bibitem[Kirschner et~al., 2019]{Kirschner:LineBO:2019}
Kirschner, J., Mutny, M., Hiller, N., Ischebeck, R., and Krause, A. (2019).
\newblock Adaptive and safe {B}ayesian optimization in high dimensions via one-dimensional subspaces.
\newblock In Chaudhuri, K. and Salakhutdinov, R., editors, {\em Proceedings of the 36th International Conference on Machine Learning}, volume~97 of {\em Proceedings of Machine Learning Research}, pages 3429--3438. PMLR.

\bibitem[Kong et~al., 2024]{Kong:DSBO:2024}
Kong, D., Huang, Y., Xie, J., Honig, E., Xu, M., Xue, S., Lin, P., Zhou, S., Zhong, S., Zheng, N., and Wu, Y.~N. (2024).
\newblock Dual-space optimization: Improved molecule sequence design by latent prompt transformer.

\bibitem[Krenn et~al., 2020]{Krenn:SELFIES:2020}
Krenn, M., Häse, F., Nigam, A., Friederich, P., and Aspuru-Guzik, A. (2020).
\newblock Self-referencing embedded strings (selfies): A 100\% robust molecular string representation.
\newblock {\em Machine Learning: Science and Technology}, 1(4):045024.

\bibitem[Kristiadi et~al., 2024]{Kristiadi:LLM:2024}
Kristiadi, A., Strieth-Kalthoff, F., Skreta, M., Poupart, P., Aspuru-Guzik, A., and Pleiss, G. (2024).
\newblock A sober look at {LLM}s for material discovery: Are they actually good for {B}ayesian optimization over molecules?
\newblock In Salakhutdinov, R., Kolter, Z., Heller, K., Weller, A., Oliver, N., Scarlett, J., and Berkenkamp, F., editors, {\em Proceedings of the 41st International Conference on Machine Learning}, volume 235 of {\em Proceedings of Machine Learning Research}, pages 25603--25622. PMLR.

\bibitem[Kusner et~al., 2017]{Kusner:GrammarVAE:2017}
Kusner, M.~J., Paige, B., and Hern{\'a}ndez-Lobato, J.~M. (2017).
\newblock Grammar variational autoencoder.
\newblock In Precup, D. and Teh, Y.~W., editors, {\em Proceedings of the 34th International Conference on Machine Learning}, volume~70 of {\em Proceedings of Machine Learning Research}, pages 1945--1954.

\bibitem[Lee et~al., 2023]{Lee:CoBo:2023}
Lee, S., Chu, J., Kim, S., Ko, J., and Kim, H.~J. (2023).
\newblock Advancing bayesian optimization via learning correlated latent space.
\newblock In Oh, A., Naumann, T., Globerson, A., Saenko, K., Hardt, M., and Levine, S., editors, {\em Advances in Neural Information Processing Systems}, volume~36, pages 48906--48917. Curran Associates, Inc.

\bibitem[Letham et~al., 2020]{Letham:ALEBO:2020}
Letham, B., Calandra, R., Rai, A., and Bakshy, E. (2020).
\newblock Re-examining linear embeddings for high-dimensional bayesian optimization.
\newblock In Larochelle, H., Ranzato, M., Hadsell, R., Balcan, M., and Lin, H., editors, {\em Advances in Neural Information Processing Systems}, volume~33, pages 1546--1558. Curran Associates, Inc.

\bibitem[Li et~al., 2018]{Li:HDBODropout:2018}
Li, C., Gupta, S., Rana, S., Nguyen, V., Venkatesh, S., and Shilton, A. (2018).
\newblock High dimensional bayesian optimization using dropout.

\bibitem[Li et~al., 2016]{Li:RPP:2016}
Li, C.-L., Kandasamy, K., Poczos, B., and Schneider, J. (2016).
\newblock High dimensional bayesian optimization via restricted projection pursuit models.
\newblock In Gretton, A. and Robert, C.~C., editors, {\em Proceedings of the 19th International Conference on Artificial Intelligence and Statistics}, volume~51 of {\em Proceedings of Machine Learning Research}, pages 884--892, Cadiz, Spain. PMLR.

\bibitem[Lodhi et~al., 2000]{Lodhi:SSKs:2000}
Lodhi, H., Shawe-Taylor, J., Cristianini, N., and Watkins, C. (2000).
\newblock Text classification using string kernels.
\newblock In Leen, T., Dietterich, T., and Tresp, V., editors, {\em Advances in Neural Information Processing Systems}, volume~13. MIT Press.

\bibitem[Loeffler et~al., 2024]{Loeffler:REINVENT:2024}
Loeffler, H.~H., He, J., Tibo, A., Janet, J.~P., Voronov, A., Mervin, L.~H., and Engkvist, O. (2024).
\newblock Reinvent 4: Modern ai--driven generative molecule design.
\newblock {\em Journal of Cheminformatics}, 16(1):20.

\bibitem[Lu et~al., 2018]{Lu:SGVAE:2018}
Lu, X., Gonzalez, J., Dai, Z., and Lawrence, N. (2018).
\newblock Structured variationally auto-encoded optimization.
\newblock In Dy, J. and Krause, A., editors, {\em Proceedings of the 35th International Conference on Machine Learning}, volume~80 of {\em Proceedings of Machine Learning Research}, pages 3267--3275. PMLR.

\bibitem[Matthews et~al., 2017]{Matthews2017gpflow}
Matthews, A. G. d.~G., {van der Wilk}, M., Nickson, T., Fujii, K., {Boukouvalas}, A., {Le{\'o}n-Villagr{\'a}}, P., Ghahramani, Z., and Hensman, J. (2017).
\newblock {{GP}flow: A {G}aussian process library using {T}ensor{F}low}.
\newblock {\em Journal of Machine Learning Research}, 18(40):1--6.

\bibitem[Maus et~al., 2022]{Maus:LOLBO:2022}
Maus, N., Jones, H., Moore, J., Kusner, M.~J., Bradshaw, J., and Gardner, J. (2022).
\newblock Local latent space bayesian optimization over structured inputs.
\newblock In Koyejo, S., Mohamed, S., Agarwal, A., Belgrave, D., Cho, K., and Oh, A., editors, {\em Advances in Neural Information Processing Systems}, volume~35, pages 34505--34518. Curran Associates, Inc.

\bibitem[Maus et~al., 2023]{Maus:ROBOT:2023}
Maus, N., Wu, K., Eriksson, D., and Gardner, J. (2023).
\newblock Discovering many diverse solutions with bayesian optimization.

\bibitem[Michael et~al., 2024]{Michael:COREL:2024}
Michael, R., Bartels, S., González-Duque, M., Zainchkovskyy, Y., Frellsen, J., Hauberg, S., and Boomsma, W. (2024).
\newblock A continuous relaxation for discrete bayesian optimization.

\bibitem[Moriconi et~al., 2020a]{Moriconi:MGPC:2020}
Moriconi, R., Deisenroth, M.~P., and Sesh~Kumar, K. (2020a).
\newblock High-dimensional bayesian optimization using low-dimensional feature spaces.
\newblock {\em Machine Learning}, 109:1925--1943.

\bibitem[Moriconi et~al., 2020b]{Moriconi:QuantGPBO:2020}
Moriconi, R., Kumar, K.~S., and Deisenroth, M.~P. (2020b).
\newblock High-dimensional bayesian optimization with projections using quantile gaussian processes.
\newblock {\em Optimization Letters}, 14:51--64.

\bibitem[Moss et~al., 2020]{Moss:BOSS:2020}
Moss, H., Leslie, D., Beck, D., Gonz\'{a}lez, J., and Rayson, P. (2020).
\newblock Boss: Bayesian optimization over string spaces.
\newblock In Larochelle, H., Ranzato, M., Hadsell, R., Balcan, M., and Lin, H., editors, {\em Advances in Neural Information Processing Systems}, volume~33, pages 15476--15486. Curran Associates, Inc.

\bibitem[Močkus, 1975]{Mockus:OriginalBO:1975}
Močkus, J. (1975).
\newblock On bayesian methods for seeking the extremum.
\newblock In Marchuk, G.~I., editor, {\em Optimization Techniques IFIP Technical Conference Novosibirsk, July 1–7, 1974}, page 400–404, Berlin, Heidelberg. Springer.

\bibitem[M\"{u}ller et~al., 2021]{Muller:LPSBO:2021}
M\"{u}ller, S., von Rohr, A., and Trimpe, S. (2021).
\newblock Local policy search with bayesian optimization.
\newblock In Ranzato, M., Beygelzimer, A., Dauphin, Y., Liang, P., and Vaughan, J.~W., editors, {\em Advances in Neural Information Processing Systems}, volume~34, pages 20708--20720. Curran Associates, Inc.

\bibitem[Mutny and Krause, 2018]{Mutny:HDBOQFF:2018}
Mutny, M. and Krause, A. (2018).
\newblock Efficient high dimensional bayesian optimization with additivity and quadrature fourier features.
\newblock In Bengio, S., Wallach, H., Larochelle, H., Grauman, K., Cesa-Bianchi, N., and Garnett, R., editors, {\em Advances in Neural Information Processing Systems}, volume~31. Curran Associates, Inc.

\bibitem[Nayebi et~al., 2019]{Nayebi:HESBO:2019}
Nayebi, A., Munteanu, A., and Poloczek, M. (2019).
\newblock A framework for {B}ayesian optimization in embedded subspaces.
\newblock In Chaudhuri, K. and Salakhutdinov, R., editors, {\em Proceedings of the 36th International Conference on Machine Learning}, volume~97 of {\em Proceedings of Machine Learning Research}, pages 4752--4761. PMLR.

\bibitem[Needleman and Wunsch, 1970]{Needleman:AminoAcids:1970}
Needleman, S.~B. and Wunsch, C.~D. (1970).
\newblock A general method applicable to the search for similarities in the amino acid sequence of two proteins.
\newblock {\em Journal of Molecular Biology}, 48(3):443--453.

\bibitem[Ngo et~al., 2024]{Ngo:CMABO:2024}
Ngo, L., Ha, H., Chan, J., Nguyen, V., and Zhang, H. (2024).
\newblock High-dimensional bayesian optimization via covariance matrix adaptation strategy.

\bibitem[Nguyen et~al., 2022]{Nguyen:MPD:2022}
Nguyen, Q., Wu, K., Gardner, J., and Garnett, R. (2022).
\newblock Local bayesian optimization via maximizing probability of descent.
\newblock In Koyejo, S., Mohamed, S., Agarwal, A., Belgrave, D., Cho, K., and Oh, A., editors, {\em Advances in Neural Information Processing Systems}, volume~35, pages 13190--13202. Curran Associates, Inc.

\bibitem[Notin et~al., 2021]{Notin:Uncert:2021}
Notin, P., Hern\'{a}ndez-Lobato, J.~M., and Gal, Y. (2021).
\newblock Improving black-box optimization in vae latent space using decoder uncertainty.
\newblock In Ranzato, M., Beygelzimer, A., Dauphin, Y., Liang, P., and Vaughan, J.~W., editors, {\em Advances in Neural Information Processing Systems}, volume~34, pages 802--814. Curran Associates, Inc.

\bibitem[Notin et~al., 2023]{Notin:ProtGym:2023}
Notin, P., Kollasch, A., Ritter, D., van Niekerk, L., Paul, S., Spinner, H., Rollins, N., Shaw, A., Orenbuch, R., Weitzman, R., Frazer, J., Dias, M., Franceschi, D., Gal, Y., and Marks, D. (2023).
\newblock Proteingym: Large-scale benchmarks for protein fitness prediction and design.
\newblock In Oh, A., Naumann, T., Globerson, A., Saenko, K., Hardt, M., and Levine, S., editors, {\em Advances in Neural Information Processing Systems}, volume~36, pages 64331--64379. Curran Associates, Inc.

\bibitem[Oh et~al., 2018]{Oh:BOCK:2018}
Oh, C., Gavves, E., and Welling, M. (2018).
\newblock {BOCK} : {B}ayesian optimization with cylindrical kernels.
\newblock In Dy, J. and Krause, A., editors, {\em Proceedings of the 35th International Conference on Machine Learning}, volume~80 of {\em Proceedings of Machine Learning Research}, pages 3868--3877. PMLR.

\bibitem[Oh et~al., 2019]{Oh:COMBO:2019}
Oh, C., Tomczak, J., Gavves, E., and Welling, M. (2019).
\newblock Combinatorial bayesian optimization using the graph cartesian product.
\newblock In Wallach, H., Larochelle, H., Beygelzimer, A., d\textquotesingle Alch\'{e}-Buc, F., Fox, E., and Garnett, R., editors, {\em Advances in Neural Information Processing Systems}, volume~32. Curran Associates, Inc.

\bibitem[Palar and Shimoyama, 2017]{Palar:ASM:2017}
Palar, P.~S. and Shimoyama, K. (2017).
\newblock Exploiting active subspaces in global optimization: how complex is your problem?
\newblock In {\em Proceedings of the Genetic and Evolutionary Computation Conference Companion}, GECCO '17, page 1487–1494, New York, NY, USA. Association for Computing Machinery.

\bibitem[Papenmeier et~al., 2022]{Papenmeier:BAxUS:2022}
Papenmeier, L., Nardi, L., and Poloczek, M. (2022).
\newblock Increasing the scope as you learn: Adaptive bayesian optimization in nested subspaces.
\newblock In Koyejo, S., Mohamed, S., Agarwal, A., Belgrave, D., Cho, K., and Oh, A., editors, {\em Advances in Neural Information Processing Systems}, volume~35, pages 11586--11601. Curran Associates, Inc.

\bibitem[Papenmeier et~al., 2024]{Papenmeier:BOUNCE:2024}
Papenmeier, L., Nardi, L., and Poloczek, M. (2024).
\newblock Bounce: Reliable high-dimensional bayesian optimization for combinatorial and mixed spaces.

\bibitem[Pedrielli and Ng, 2016]{Pedrielli:GSTAR:2016}
Pedrielli, G. and Ng, S.~H. (2016).
\newblock G-star: A new kriging-based trust region method for global optimization.
\newblock In {\em 2016 Winter Simulation Conference (WSC)}, pages 803--814.

\bibitem[Penner, 2022]{Penner:ProtMODULI:2022}
Penner, R. (2022).
\newblock Protein geometry, function and mutation.

\bibitem[Penubothula et~al., 2021]{Penubothula:FirstOrder:2021}
Penubothula, S., Kamanchi, C., and Bhatnagar, S. (2021).
\newblock Novel first order bayesian optimization with an application to reinforcement learning.
\newblock {\em Applied Intelligence}, 51(3):1565--1579.

\bibitem[Picheny et~al., 2023]{Picheny2023trieste}
Picheny, V., Berkeley, J., Moss, H.~B., Stojic, H., Granta, U., Ober, S.~W., Artemev, A., Ghani, K., Goodall, A., Paleyes, A., Vakili, S., Pascual-Diaz, S., Markou, S., Qing, J., Loka, N. R. B.~S., and Couckuyt, I. (2023).
\newblock Trieste: Efficiently exploring the depths of black-box functions with tensorflow.

\bibitem[Pyzer-Knapp, 2018]{Pyzer-Knapp:BODrugDiscovery:2018}
Pyzer-Knapp, E.~O. (2018).
\newblock Bayesian optimization for accelerated drug discovery.
\newblock {\em IBM Journal of Research and Development}, 62(6):2:1--2:7.

\bibitem[Qian et~al., 2016]{Qian:SREIMGPO:2016}
Qian, H., Hu, Y.-Q., and Yu, Y. (2016).
\newblock Derivative-free optimization of high-dimensional non-convex functions by sequential random embeddings.
\newblock In {\em IJCAI}, pages 1946--1952.

\bibitem[Rana et~al., 2017]{Santu:ElasticGPs:2017}
Rana, S., Li, C., Gupta, S., Nguyen, V., and Venkatesh, S. (2017).
\newblock High dimensional {B}ayesian optimization with elastic {G}aussian process.
\newblock In Precup, D. and Teh, Y.~W., editors, {\em Proceedings of the 34th International Conference on Machine Learning}, volume~70 of {\em Proceedings of Machine Learning Research}, pages 2883--2891. PMLR.

\bibitem[Raponi et~al., 2020]{Raponi:PCABO:2020}
Raponi, E., Wang, H., Bujny, M., Boria, S., and Doerr, C. (2020).
\newblock High dimensional bayesian optimization assisted by principal component analysis.
\newblock In {\em Parallel Problem Solving from Nature--PPSN XVI: 16th International Conference, PPSN 2020, Leiden, The Netherlands, September 5-9, 2020, Proceedings, Part I 16}, pages 169--183. Springer.

\bibitem[Rasmussen and Williams, 2006]{RasmussenWilliams:GPs:2006}
Rasmussen, C.~E. and Williams, C. K.~I. (2006).
\newblock {\em Gaussian processes for machine learning}.
\newblock Adaptive computation and machine learning. MIT Press, Cambridge, Mass.

\bibitem[Regis, 2016]{Rommel:TRIKE:2016}
Regis, R.~G. (2016).
\newblock Trust regions in kriging-based optimization with expected improvement.
\newblock {\em Engineering Optimization}, 48(6):1037--1059.

\bibitem[Rezende et~al., 2014]{Rezende:VAE:2014}
Rezende, D.~J., Mohamed, S., and Wierstra, D. (2014).
\newblock Stochastic backpropagation and approximate inference in deep generative models.
\newblock In Xing, E.~P. and Jebara, T., editors, {\em Proceedings of the 31st International Conference on Machine Learning}, volume~32 of {\em Proceedings of Machine Learning Research}, pages 1278--1286, Bejing, China. PMLR.

\bibitem[Rolland et~al., 2018]{Rolland:GAddGPUCB:2018}
Rolland, P., Scarlett, J., Bogunovic, I., and Cevher, V. (2018).
\newblock High-dimensional bayesian optimization via additive models with overlapping groups.
\newblock In Storkey, A. and Perez-Cruz, F., editors, {\em Proceedings of the Twenty-First International Conference on Artificial Intelligence and Statistics}, volume~84 of {\em Proceedings of Machine Learning Research}, pages 298--307. PMLR.

\bibitem[Ru et~al., 2020]{Ru:CoCaBO:2020}
Ru, B., Alvi, A., Nguyen, V., Osborne, M.~A., and Roberts, S. (2020).
\newblock {B}ayesian optimisation over multiple continuous and categorical inputs.
\newblock In III, H.~D. and Singh, A., editors, {\em Proceedings of the 37th International Conference on Machine Learning}, volume 119 of {\em Proceedings of Machine Learning Research}, pages 8276--8285. PMLR.

\bibitem[Santoni et~al., 2023]{SantoniDoerr:HDBO:2023}
Santoni, M.~L., Raponi, E., Leone, R.~D., and Doerr, C. (2023).
\newblock Comparison of high-dimensional bayesian optimization algorithms on bbob.

\bibitem[Shahriari et~al., 2016]{Shahriari2016BOreview}
Shahriari, B., Swersky, K., Wang, Z., Adams, R.~P., and de~Freitas, N. (2016).
\newblock Taking the human out of the loop: A review of bayesian optimization.
\newblock {\em Proceedings of the IEEE}, 104(1):148--175.

\bibitem[Shekhar and Javidi, 2021]{Shekhar:AlgFOO:2021}
Shekhar, S. and Javidi, T. (2021).
\newblock Significance of gradient information in bayesian optimization.
\newblock In Banerjee, A. and Fukumizu, K., editors, {\em Proceedings of The 24th International Conference on Artificial Intelligence and Statistics}, volume 130 of {\em Proceedings of Machine Learning Research}, pages 2836--2844. PMLR.

\bibitem[Shervashidze et~al., 2011]{Shervashidze:WLGraphKernel:2011}
Shervashidze, N., Schweitzer, P., Van~Leeuwen, E.~J., Mehlhorn, K., and Borgwardt, K.~M. (2011).
\newblock Weisfeiler-lehman graph kernels.
\newblock {\em Journal of Machine Learning Research}, 12(9).

\bibitem[Shmakov et~al., 2023]{Shmakov:RTDKBO:2023}
Shmakov, A., Naug, A., Gundecha, V., Ghorbanpour, S., Gutierrez, R.~L., Babu, A.~R., Guillen, A., and Sarkar, S. (2023).
\newblock Rtdk-bo: High dimensional bayesian optimization with reinforced transformer deep kernels.
\newblock In {\em 2023 IEEE 19th International Conference on Automation Science and Engineering (CASE)}, pages 1--8.

\bibitem[Snoek et~al., 2012]{Snoek:PracticalBO:2012}
Snoek, J., Larochelle, H., and Adams, R.~P. (2012).
\newblock Practical bayesian optimization of machine learning algorithms.
\newblock In Pereira, F., Burges, C., Bottou, L., and Weinberger, K., editors, {\em Advances in Neural Information Processing Systems}, volume~25. Curran Associates, Inc.

\bibitem[Song et~al., 2022]{Song:MTCTS:2022}
Song, L., Xue, K., Huang, X., and Qian, C. (2022).
\newblock Monte carlo tree search based variable selection for high dimensional bayesian optimization.
\newblock In Koyejo, S., Mohamed, S., Agarwal, A., Belgrave, D., Cho, K., and Oh, A., editors, {\em Advances in Neural Information Processing Systems}, volume~35, pages 28488--28501. Curran Associates, Inc.

\bibitem[Springenberg et~al., 2016]{Springberg:BOHEMIANN:2016}
Springenberg, J.~T., Klein, A., Falkner, S., and Hutter, F. (2016).
\newblock Bayesian optimization with robust bayesian neural networks.
\newblock In Lee, D., Sugiyama, M., Luxburg, U., Guyon, I., and Garnett, R., editors, {\em Advances in Neural Information Processing Systems}, volume~29. Curran Associates, Inc.

\bibitem[Srinivas et~al., 2012]{Srinivas:UCB:2012}
Srinivas, N., Krause, A., Kakade, S.~M., and Seeger, M.~W. (2012).
\newblock Information-theoretic regret bounds for gaussian process optimization in the bandit setting.
\newblock {\em IEEE Transactions on Information Theory}, 58(5):3250--3265.

\bibitem[Stanton et~al., 2024]{Stanton:Ehrlich:2024}
Stanton, S., Alberstein, R., Frey, N., Watkins, A., and Cho, K. (2024).
\newblock Closed-form test functions for biophysical sequence optimization algorithms.

\bibitem[Stanton et~al., 2022]{Stanton:LAMBO:2022}
Stanton, S., Maddox, W., Gruver, N., Maffettone, P., Delaney, E., Greenside, P., and Wilson, A.~G. (2022).
\newblock Accelerating {B}ayesian optimization for biological sequence design with denoising autoencoders.
\newblock In Chaudhuri, K., Jegelka, S., Song, L., Szepesvari, C., Niu, G., and Sabato, S., editors, {\em Proceedings of the 39th International Conference on Machine Learning}, volume 162 of {\em Proceedings of Machine Learning Research}, pages 20459--20478. PMLR.

\bibitem[Swersky et~al., 2020]{Swersky:Amortized:2020}
Swersky, K., Rubanova, Y., Dohan, D., and Murphy, K. (2020).
\newblock Amortized bayesian optimization over discrete spaces.
\newblock In Peters, J. and Sontag, D., editors, {\em Proceedings of the 36th Conference on Uncertainty in Artificial Intelligence (UAI)}, volume 124 of {\em Proceedings of Machine Learning Research}, pages 769--778. PMLR.

\bibitem[Tripp et~al., 2020]{Tripp:WeightedRetraining:2020}
Tripp, A., Daxberger, E., and Hern\'{a}ndez-Lobato, J.~M. (2020).
\newblock Sample-efficient optimization in the latent space of deep generative models via weighted retraining.
\newblock In Larochelle, H., Ranzato, M., Hadsell, R., Balcan, M., and Lin, H., editors, {\em Advances in Neural Information Processing Systems}, volume~33, pages 11259--11272. Curran Associates, Inc.

\bibitem[Tripp and Hernández-Lobato, 2023]{Tripp:MolGA:2023}
Tripp, A. and Hernández-Lobato, J.~M. (2023).
\newblock Genetic algorithms are strong baselines for molecule generation.

\bibitem[Turner et~al., 2021]{Turner:BOHyperTuning:2020}
Turner, R., Eriksson, D., McCourt, M., Kiili, J., Laaksonen, E., Xu, Z., and Guyon, I. (2021).
\newblock Bayesian optimization is superior to random search for machine learning hyperparameter tuning: Analysis of the black-box optimization challenge 2020.
\newblock In Escalante, H.~J. and Hofmann, K., editors, {\em Proceedings of the NeurIPS 2020 Competition and Demonstration Track}, volume 133 of {\em Proceedings of Machine Learning Research}, pages 3--26. PMLR.

\bibitem[Ulmasov et~al., 2016]{Ulmasov:DSA:2016}
Ulmasov, D., Baroukh, C., Chachuat, B., Deisenroth, M.~P., and Misener, R. (2016).
\newblock Bayesian optimization with dimension scheduling: Application to biological systems.
\newblock In Kravanja, Z. and Bogataj, M., editors, {\em 26th European Symposium on Computer Aided Process Engineering}, volume~38 of {\em Computer Aided Chemical Engineering}, pages 1051--1056. Elsevier.

\bibitem[Urbina et~al., 2022]{Urbina:DualUse:2022}
Urbina, F., Lentzos, F., Invernizzi, C., and Ekins, S. (2022).
\newblock Dual use of artificial-intelligence-powered drug discovery.
\newblock {\em Nature Machine Intelligence}, 4(3):189–191.

\bibitem[{van der Wilk} et~al., 2020]{Wilk2020gpflow}
{van der Wilk}, M., Dutordoir, V., John, S., Artemev, A., Adam, V., and Hensman, J. (2020).
\newblock A framework for interdomain and multioutput {G}aussian processes.
\newblock {\em arXiv:2003.01115}.

\bibitem[Verma et~al., 2022]{Verma:HDinvariance:2022}
Verma, E., Chakraborty, S., and Griffiths, R.-R. (2022).
\newblock Highdimensional bayesian optimization with invariance.
\newblock In {\em ICML Workshop on Adaptive Experimental Design and Active Learning}.

\bibitem[Wan et~al., 2021]{Wan:CASMOPOLITAN:2021}
Wan, X., Nguyen, V., Ha, H., Ru, B., Lu, C., and Osborne, M.~A. (2021).
\newblock Think global and act local: Bayesian optimisation over high-dimensional categorical and mixed search spaces.
\newblock In Meila, M. and Zhang, T., editors, {\em Proceedings of the 38th International Conference on Machine Learning}, volume 139 of {\em Proceedings of Machine Learning Research}, pages 10663--10674. PMLR.

\bibitem[Wang et~al., 2023]{Wang:BOAdvances:2023}
Wang, X., Jin, Y., Schmitt, S., and Olhofer, M. (2023).
\newblock Recent advances in bayesian optimization.
\newblock {\em ACM Comput. Surv.}, 55(13s).

\bibitem[Wang et~al., 2018]{Wang:BatchedEnsembleHDBO:2018}
Wang, Z., Gehring, C., Kohli, P., and Jegelka, S. (2018).
\newblock Batched large-scale bayesian optimization in high-dimensional spaces.
\newblock In Storkey, A. and Perez-Cruz, F., editors, {\em Proceedings of the Twenty-First International Conference on Artificial Intelligence and Statistics}, volume~84 of {\em Proceedings of Machine Learning Research}, pages 745--754. PMLR.

\bibitem[Wang et~al., 2016]{Wang:REMBO:2016}
Wang, Z., Hutter, F., Zoghi, M., Matheson, D., and De~Feitas, N. (2016).
\newblock Bayesian optimization in a billion dimensions via random embeddings.
\newblock {\em Journal of Artificial Intelligence Research}, 55:361--387.

\bibitem[Wang et~al., 2013]{Wang2013rembo}
Wang, Z., Zoghi, M., Hutter, F., Matheson, D., and De~Freitas, N. (2013).
\newblock Bayesian optimization in high dimensions via random embeddings.
\newblock In {\em Proceedings of the Twenty-Third International Joint Conference on Artificial Intelligence}, IJCAI '13, page 1778–1784.

\bibitem[Weininger, 1988]{Weininger:SMILES:1988}
Weininger, D. (1988).
\newblock Smiles, a chemical language and information system. 1. introduction to methodology and encoding rules.
\newblock {\em J. Chem. Inf. Comput. Sci.}, 28(1):31--36.

\bibitem[Wilkins, 1668]{Wilkins:PhilLang:1668}
Wilkins, J. (1668).
\newblock {\em An Essay towards a Real Character And a Philosophical Language}.
\newblock Royal Society, London.

\bibitem[Winkel et~al., 2021]{Munir:SOLID:2021}
Winkel, M.~A., Stallrich, J.~W., Storlie, C.~B., and Reich, B.~J. (2021).
\newblock Sequential optimization in locally important dimensions.
\newblock {\em Technometrics}, 63(2):236--248.

\bibitem[Wu et~al., 2017]{Wu:Gradients:2017}
Wu, J., Poloczek, M., Wilson, A.~G., and Frazier, P. (2017).
\newblock Bayesian optimization with gradients.
\newblock In Guyon, I., Luxburg, U.~V., Bengio, S., Wallach, H., Fergus, R., Vishwanathan, S., and Garnett, R., editors, {\em Advances in Neural Information Processing Systems}, volume~30. Curran Associates, Inc.

\bibitem[Wycoff et~al., 2021]{Wycoff:AS:2021}
Wycoff, N., Binois, M., and Wild, S.~M. (2021).
\newblock Sequential learning of active subspaces.
\newblock {\em Journal of Computational and Graphical Statistics}, 30(4):1224--1237.

\bibitem[Xu and Zhe, 2024]{Xu:StdVanillaBO:2024}
Xu, Z. and Zhe, S. (2024).
\newblock Standard gaussian process can be excellent for high-dimensional bayesian optimization.

\bibitem[Yenicelik, 2020]{Yenicelik:BORING:2020}
Yenicelik, D. (2020).
\newblock Parameter optimization using high-dimensional bayesian optimization.

\bibitem[Yin et~al., 2024]{Yin:TSBO:2024}
Yin, Y., Wang, Y., and Li, P. (2024).
\newblock High-dimensional bayesian optimization via semi-supervised learning with optimized unlabeled data sampling.

\bibitem[Zhan, 2024]{Zhan:ECI:2024}
Zhan, D. (2024).
\newblock Expected coordinate improvement for high-dimensional bayesian optimization.

\bibitem[Zhang et~al., 2019]{Zhang:SIR:2019}
Zhang, M., Li, H., and Su, S. (2019).
\newblock High dimensional bayesian optimization via supervised dimension reduction.

\bibitem[Zhao et~al., 2024]{Zhao:AIBO:2024}
Zhao, J., Yang, R., Qiu, S., and Wang, Z. (2024).
\newblock Unleashing the potential of acquisition functions in high-dimensional bayesian optimization.

\bibitem[Zhou et~al., 2021]{Zhou:TRPBO:2021}
Zhou, J., Yang, Z., Si, Y., Kang, L., Li, H., Wang, M., and Zhang, Z. (2021).
\newblock A trust-region parallel bayesian optimization method for simulation-driven antenna design.
\newblock {\em IEEE Transactions on Antennas and Propagation}, 69(7):3966--3981.

\bibitem[Zhu et~al., 2022]{Zhu:TorchDrug:2022}
Zhu, Z., Shi, C., Zhang, Z., Liu, S., Xu, M., Yuan, X., Zhang, Y., Chen, J., Cai, H., Lu, J., Ma, C., Liu, R., Xhonneux, L.-P., Qu, M., and Tang, J. (2022).
\newblock Torchdrug: A powerful and flexible machine learning platform for drug discovery.

\bibitem[Ziomek and Bou~Ammar, 2023]{Ziomek:RDUCB:2023}
Ziomek, J.~K. and Bou~Ammar, H. (2023).
\newblock Are random decompositions all we need in high dimensional {B}ayesian optimisation?
\newblock In Krause, A., Brunskill, E., Cho, K., Engelhardt, B., Sabato, S., and Scarlett, J., editors, {\em Proceedings of the 40th International Conference on Machine Learning}, volume 202 of {\em Proceedings of Machine Learning Research}, pages 43347--43368. PMLR.

\end{thebibliography}

\newpage
\appendix

\section{Appendix}

\subsection{Methods Overview}
\begin{longtable}{| p{.2\textwidth} | p{.2\textwidth} | p{.5\textwidth} | p{.1\textwidth} |}
\toprule
\textbf{Method} & \textbf{Date} \hspace{10em}(first occurrence) & \textbf{Reference} & \textbf{Code Available} \\
\midrule
SOLID & January 23, 2019 & \cite{Munir:SOLID:2021} & \ding{55} \\
Deep GPs & May 7, 2019 & \cite{Hebbal:DeepGPs:2021} & \ding{55} \\
ASM & July 15, 2017 & \cite{Palar:ASM:2017}  & \ding{55} \\
Add-GP-UCB & May 13, 2015 & \cite{Han:AddGPUCB:2021} & \ding{51} \\
TuRBO & December 8, 2019 & \cite{Eriksson:TuRBO:2019} & \ding{51} \\
LOL-BO & January 28, 2022 & \cite{Maus:LOLBO:2022} & \ding{51} \\
ROBOT & October 20, 2022 & \cite{Maus:ROBOT:2023} & \ding{51} \\
REMBO & January 9, 2013 & \cite{Wang:REMBO:2016} & \ding{51} \\
SAASBO & June 10, 2021 & \cite{Eriksson:SAASBO:2021} & \ding{51} \\
Dropout & August 19, 2017 & \cite{Li:HDBODropout:2018} & \ding{55} \\
BAxUS & November 28, 2022 & \cite{Papenmeier:BAxUS:2022} & \ding{51} \\
LineBO & June 10, 2019 & \cite{Kirschner:LineBO:2019} & \ding{51} \\
ALEBO & December 6, 2020 & \cite{Letham:ALEBO:2020} & \ding{51} \\
HeSBO & May 24, 2019 & \cite{Nayebi:HESBO:2019} & \ding{51} \\
BORING & October 5, 2020 & \cite{Yenicelik:BORING:2020} & \ding{55} \\
REMBO 2.0 & October 18, 2019 & \cite{Binois:REMBO2:2020} & \ding{55} \\
Quantile-GP BO & February 1, 2020 & \cite{Moriconi:QuantGPBO:2020} & \ding{55} \\
Warped REMBO & January 1, 2015 & \cite{Binois:WarpedREMBO:2015} & \ding{55} \\
{\tiny closed-form} ASM & September 22, 2020 & \cite{Wycoff:AS:2021} & \ding{51} \\
Active manifolds & May 24, 2019 & \cite{Bridges:ActiveManifolds:2019} & \ding{55} \\
Deep GPs (MO) & January 1, 2019 & \cite{Hebbal:DeepGPMO:2019} & \ding{55} \\
LADDER & December 6, 2021 & \cite{Deshwal:LADDER:2021} & \ding{51} \\
Attr. Adjustment & August 6, 2018 & \cite{Eissman:AttrAdjust:2018} & \ding{55} \\
VAEs DML & November 1, 2021 & \cite{Grosnit:VAEDML:2021} & \ding{51} \\
LSBO & February 28, 2018 & \cite{Bombarelli:AutoChemDesign:2018} & \ding{51} \\
Weigh. Retraining & October 25, 2020 & \cite{Tripp:WeightedRetraining:2020} & \ding{51} \\
MORBO (MO) & September 22, 2021 & \cite{Daulton:MORBO:2022} & \ding{51} \\
TREGO & October 10, 2022 & \cite{Diouane:TREGO:2023} & \ding{55} \\
TRIKE & August 8, 2015 & \cite{Rommel:TRIKE:2016}  & \ding{55} \\
TRPBO & November 21, 2020 & \cite{Zhou:TRPBO:2021} & \ding{55} \\
D-SKIP & December 3, 2018 & \cite{Eriksson:DSKIP:2018} & \ding{51} \\
RDUCB & May 29, 2023 & \cite{Ziomek:RDUCB:2023} & \ding{51} \\
G-Add-GP-UCB & April 9, 2018 & \cite{Rolland:GAddGPUCB:2018} & \ding{55} \\
QFF & December 3, 2018 & \cite{Mutny:HDBOQFF:2018} & \ding{55} \\
SI-BO & December 5, 2013 & \cite{Djolonga:HDBandits:2013} & \ding{55} \\
SRE-IMGPO & July 9, 2016 & \cite{Qian:SREIMGPO:2016} & \ding{55} \\
HDS & June 27, 2012 & \cite{Chen:HDS:2012} & \ding{55} \\
AL of LEs & October 24, 2013 & \cite{Garnett:ActiveLLEmb:2013} & \ding{51} \\
SG-VAE & July 3, 2018 & \cite{Lu:SGVAE:2018} & \ding{55} \\
BO+MCMC & April 10, 2017 & \cite{Gardner:AddStruct:2017} & \ding{51} \\
Ensemble BO & March 31, 2018 & \cite{Wang:BatchedEnsembleHDBO:2018} & \ding{51} \\
BOCK & March 3, 2018 & \cite{Oh:BOCK:2018} & \ding{51} \\
COMBO & December 8, 2019 & \cite{Oh:COMBO:2019} & \ding{51} \\
MGPC-BO & September 1, 2020 & \cite{Moriconi:MGPC:2020} & \ding{51} \\
G-STAR & December 1, 2016 & \cite{Pedrielli:GSTAR:2016} & \ding{55} \\
CASMOPOLITAN & June 18, 2021 & \cite{Wan:CASMOPOLITAN:2021} & \ding{51} \\
Vanilla BO & February 25, 2024 & \cite{Hvarfner:VanilaBO:2024} & \ding{51} \\
Bounce & July 2, 2023 &  \cite{Papenmeier:BOUNCE:2024} & \ding{51} \\
EGP & August 7, 2017 & \cite{Santu:ElasticGPs:2017} & \ding{55} \\
CoBO & December 10, 2023 & \cite{Lee:CoBo:2023} & \ding{51} \\
DSO & February 27, 2024 & \cite{Kong:DSBO:2024}  & \ding{55} \\
MPD & January 16, 2023 & \cite{Nguyen:MPD:2022} & \ding{51} \\
GIBO & November 22, 2021 & \cite{Muller:LPSBO:2021} & \ding{51} \\
PR & October 18, 2022 & \cite{Daulton:PR:2022} & \ding{51} \\
RPP & May 9, 2016 & \cite{Li:RPP:2016} & \ding{55} \\
MCTS-VS & October 31, 2022 & \cite{Song:MTCTS:2022} & \ding{51} \\
Standard BO & February 5, 2024 & \cite{Xu:StdVanillaBO:2024} & \ding{51} \\
{\tiny Scalable} FOBO & June 16, 2022 & \cite{Ament:SBOSDA:2022} & \ding{51} \\
FOBO & December 6, 2017 & \cite{Ahmed:FOBO:2016} & \ding{55} \\
BOSS & October 2, 2020 & \cite{Moss:BOSS:2020} & \ding{51} \\
SILBO & May 29, 2020 & \cite{Jingfan:SILBOHDBO:2020} & \ding{51} \\
SCoreBO & April 21, 2023 & \cite{Hvarfner:SCOREBO:2023} & \ding{51} \\
HEBO & December 7, 2020 & \cite{Cowen:HEBO:2022} & \ding{51} \\
Tree Add-GP-UCB & May 1, 2021 & \cite{Han:AddGPUCB:2021} & \ding{51} \\
SIR/SDR & July 21, 2019 & \cite{Zhang:SIR:2019} & \ding{51} \\
GaBO & November 22, 2021 & \cite{Jaquier:GABO:2020} & \ding{51} \\
ECI & April 18, 2024 & \cite{Zhan:ECI:2024} & \ding{51} \\
MAVE-BO & March 8, 2024 & \cite{Hu:MAVEBO:2024} & \ding{55} \\
CMA-BO & February 5, 2024 & \cite{Ngo:CMABO:2024} & \ding{51} \\
RTDK-BO & October 5, 2023 & \cite{Shmakov:RTDKBO:2023} & \ding{55} \\
PG-LBO & December 28, 2023 & \cite{Chen:PGLBO:2024} & \ding{51} \\
TSBO & May 4, 2023 & \cite{Yin:TSBO:2024} & \ding{55} \\
BODi & March 3, 2023 & \cite{Deshwal:BODI:2023} & \ding{51} \\
{\tiny Mahalanobis} BatchBO & November 2, 2022 & \cite{Horiguchi:MahalaBatchBO:2022} & \ding{51} \\
KPCA-BO & April 28, 2022 & \cite{Antonov:KPCABO2022} & \ding{51} \\
PCA-BO & July 2, 2020 & \cite{Raponi:PCABO:2020} & \ding{51} \\
DSA & November 18, 2015 & \cite{Ulmasov:DSA:2016} & \ding{51} \\
RPA-GP & June 12, 2020 & \cite{Delbridge:RPAGP:2020} & \ding{51} \\
HD-GaBO & December 6, 2020 & \cite{Jaquier:HDGABO:2020} & \ding{51} \\
Amortized BO & May 27, 2020 & \cite{Swersky:Amortized:2020} & \ding{55} \\
d-KG & December 4, 2017 & \cite{Wu:Gradients:2017} & \ding{51} \\
{\tiny Prabuchandran’s } FOBO & March 1, 2021 & \cite{Penubothula:FirstOrder:2021} & \ding{55} \\
AlgFOO & March 18, 2021 & \cite{Shekhar:AlgFOO:2021} & \ding{55} \\
GTBO & October 5, 2023 & \cite{Hellsten:GTBO:2023} & \ding{51} \\
CoCaBo & June 12, 2020 & \cite{Ru:CoCaBO:2020} & \ding{51} \\
MercBO & February 2, 2019 & \cite{Deshwal:MercBO:2021} & \ding{51} \\
HyBO & July 18, 2021 & \cite{Deshwal:HybridSpaces:2021} & \ding{51} \\
BOCS & July 10, 2018 & \cite{Baptista:Combinatorial:2018} & \ding{51} \\
LaMBO & July 22, 2022 & \cite{Stanton:LAMBO:2022} & \ding{51} \\
LaMBO-2 & December 12, 2023 & \cite{Gruver:LAMBO2:2023} & \ding{51} \\
G.M. \& Lobato & March 1, 2020 & \cite{GarridoMerchan:CatIntBO:2020} & \ding{55} \\
BOHAMIANN & December 5, 2016 & \cite{Springberg:BOHEMIANN:2016} & \ding{51} \\
AIBO & February 16, 2023 & \cite{Zhao:AIBO:2024} & \ding{51} \\
CoRel & April 26, 2024 & \cite{Michael:COREL:2024} & \ding{51} \\
VAE-BO+Inv & July 22, 2022 & \cite{Verma:HDinvariance:2022} & \ding{55} \\
Uncert & November 09, 2021 & \cite{Notin:Uncert:2021} & \ding{51} \\
GAUCHE & September 21, 2023 & \cite{Griffiths:GAUCHE:2023} & \ding{51} \\
LLMs & July 21, 2024 & \cite{Kristiadi:LLM:2024} & \ding{51} \\
AntBO & January 23, 2023 & \cite{Khan:AntBO:2023} & \ding{51} \\ 
RMF & May 20, 2022 & \cite{Kim:RMF:2022} & \ding{55} \\ 
\bottomrule
\caption{References for the methods presented in the taxonomy}
\label{tab:appendix:full_taxonomy}
\end{longtable}

\subsubsection{On how HDBO can be applied to discrete sequences}
\begin{figure}
    \centering
    \includegraphics[width=0.9\linewidth]{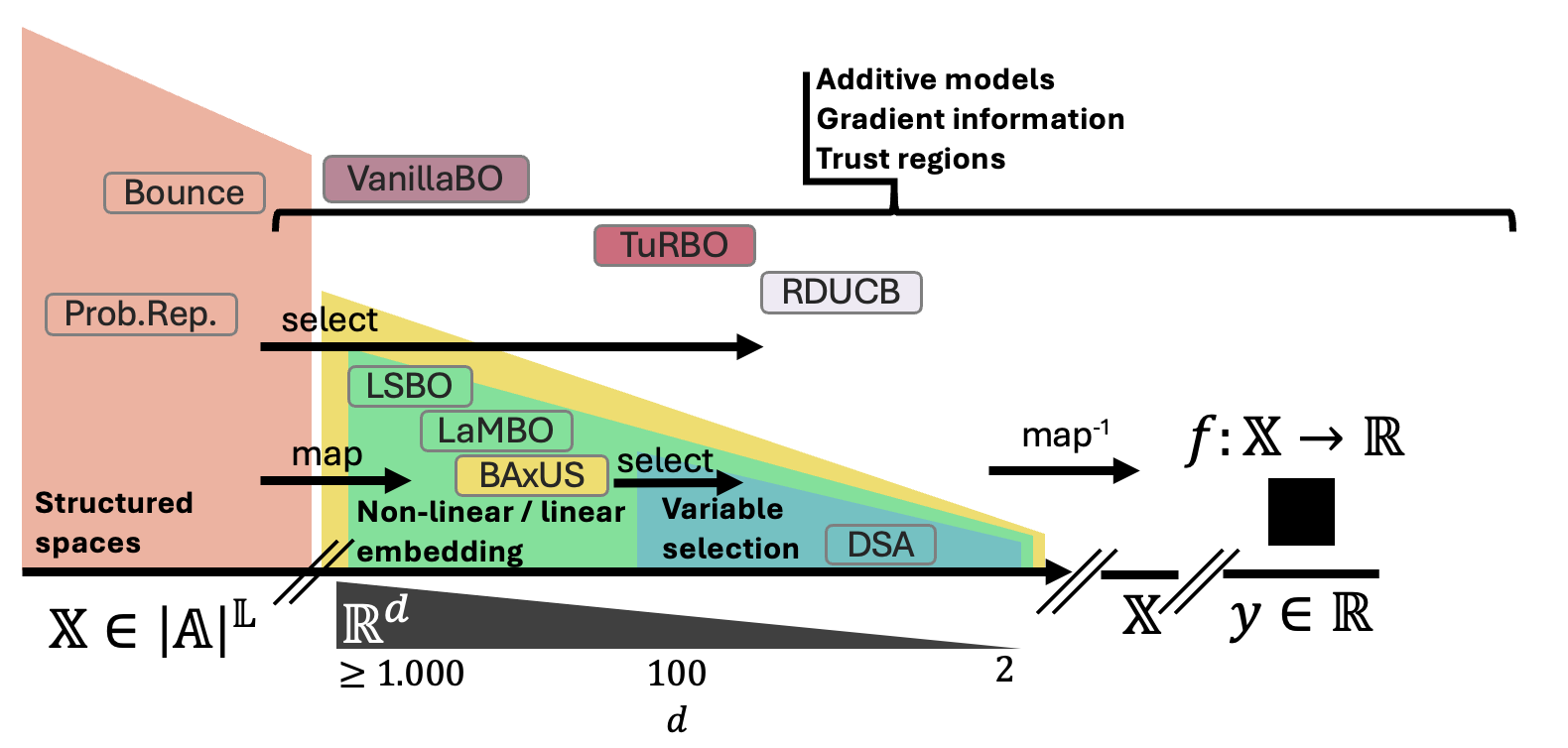}
    \caption{Overview of problem-space (x-axis) and how the categories act on the space. The black box ultimately maps from the discrete sequences of alphabet elements to a real value.
    The BO methods can act on the original (discrete) space, linear or non-linear mappings of it or selected variables of the input space or a mapping of it. These continuous versions can also accommodate one-hot representations.
    }
    \label{fig:appendix:bo_overview}
\end{figure}

Fig.~\ref{fig:appendix:bo_overview} shows an overview of the different ways in which HDBO could be applied to discrete sequences. On the left, we have the original sequence space with inputs being a list of categorical variables belonging to some alphabet; methods like Bounce, Prob. Rep and those that work on mixed/categorial inputs in the \textit{Structured Spaces} family (see Sec.~\ref{sec:taxonomy:structured-spaces}) work in this space. Since the original sequence space is large for the problems we are interested in, several other methods rely on subspaces from the original problem. Such subspaces can be constructed either as linear projections (Sec.~\ref{sec:taxonomy:linear_embeddings}) or as learned non-linear latent variables (Sec.~\ref{sec:taxonomy:non-linear-embeddings}). Once a latent continuous representation is available, all other families in the taxonomy become available and could be leveraged for latent space optimization. In these cases, there is usually a mapping between the subspace and the original high-dimensional space which allows for computing objective functions. In the linear case an up-projection is given by multiplying with a matrix of the right shape, and in the case of non-linear representations it is usually a decoding process.

In this paper, we explore applying the discrete optimizers of the \textit{Structured Spaces} family in the sequence spaces of several Ehrlich problems (Sec.~\ref{sec:benchmark:ehrlich}), as well as small molecule optimization using SELFIES representations. The continuous optimizers chosen from the other families are applied directly in one-hot space in the case of Ehrlich, or in the latent space of a Variational Autoencoder learned for SELFIES strings (Sec.~\ref{sec:benchmark:pmo}).

\subsection{Reproducing results}
\label{sec:appendix:reproducing-specific-solvers}

\paragraph{Bounce.} We forked the official open source implementation of \texttt{Bounce}\footnote{\url{https://github.com/LeoIV/bounce}} and added an interface between \texttt{poli}'s black boxes and their optimizer. Moreover, we made their implementation pip-installable. \texttt{Bounce}'s implementation is originally provided with an MIT license.

\paragraph{Probabilistic Reparametrization} provides an open-source implementation built on \texttt{GPyTorch} and \texttt{BoTorch}.\footnote{\url{https://github.com/facebookresearch/bo_pr}} They provide a \texttt{pip} installable Python package which did not install until the dependencies mentioned above were fixed (to 1.11 and 0.7 respectively); further, the environment provided had to be updated by replacing the deprecated \texttt{scikit-learn} installation in a fork of their repository. After implementing an interface for \texttt{poli} black boxes, we relied on their script \texttt{run\_one\_replication.py} to implement a custom solver. The original PR code is provided with an MIT License.

\paragraph{SAASBO, Hvarfner's Vanilla BO} were all implemented by following the tutorials in \texttt{Ax}. Ax provides models for SAASBO, and we implemented a \texttt{BoTorch} model following the original implementation. We also provide an implementation of \texttt{ALEBO} using \texttt{Ax}. Ax is provided with an MIT License, and Hvarfner's original code does not have a license in GitHub yet.

\paragraph{Turbo} was implemented by following the tutorial on BoTorch,\footnote{\url{https://github.com/pytorch/botorch/blob/main/tutorials/turbo_1.ipynb}} which can be found on GitHub under MIT License.

\paragraph{BAxUS} is implemented using the Python package provided by the authors.\footnote{\url{https://github.com/LeoIV/BAxUS}} Since their work builds on the original TuRBO code from Uber, it inherits their license.

\subsection{Training VAEs on SELFIES}
\label{sec:appendix:training-vaes-on-selfies}

\paragraph{Models.} We use \texttt{PyTorch} to implement a VAE with [one\_hot\_input, 1024, 512, 256, latent\_dim] encoder and a symmetric decoder, using ReLU activations. The latent space prior used is a standard Gaussian, and the decoded distribution is a categorical. We use \texttt{torch.distributions} to compute ELBO losses without any $\beta$ weighting.\footnote{The exact implementation can be found here: \url{https://github.com/MachineLearningLifeScience/hdbo_benchmark/blob/1a81f7fbd531eb4dc70f84828d6efc1f2ee53e5e/src/hdbo_benchmark/generative_models/vae_selfies.py}. We are currently exploring a generative model with more expressive power, and plan to update the benchmarks accordingly in our project's website.}

\paragraph{Training regimes.} We train for a maximum of 1000 epochs using early stopping with a patience of 50 epochs. Our batch sizes are $512$, learning rates are $5\times10^{-4}$, and optimizer is AdamW.\footnote{The training script can be found here: \url{https://github.com/MachineLearningLifeScience/hdbo_benchmark/blob/master/src/hdbo_benchmark/experiments/training_vae_on_zinc250k/train_vae.py}}






\subsection{Technical details on \poli{} and \polibaselines{}}
\label{sec:appendix:technical-details-on-poli-and-poli-baselines}

This section gives a short overview on the possibilities of \poli{} and \polibaselines{}. 
For further information please refer to the full online documentation under \url{https://machinelearninglifescience.github.io/poli-docs/}.
\begin{figure}
    \centering
    \includegraphics[width=\textwidth]{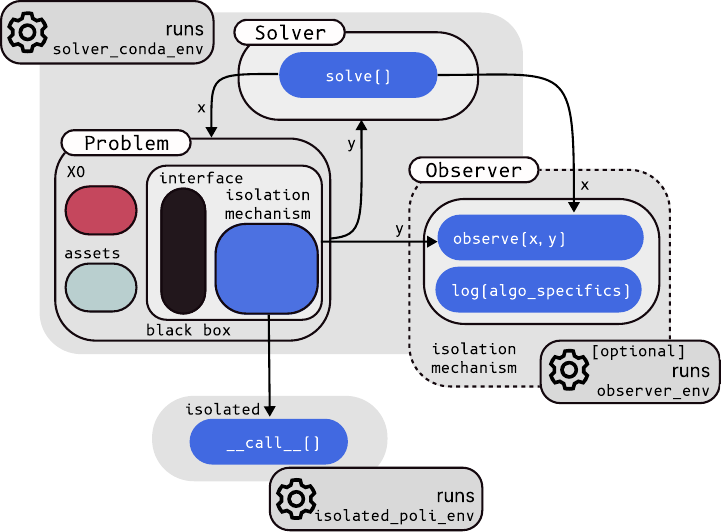}
    \caption{\texttt{poli}'s isolation process for complex environments}
    \label{fig:poli:isolation-process}
\end{figure}

The aim of \poli{} is to make it easier to benchmark new algorithms on new problems. There are three components to our framework: \textbf{problems} created by problem factories which contain black boxes, \textbf{solvers} that optimize these problems (aiming always at maximization), and \textbf{observers} which get attached to black boxes and log every single black box call. Fig.~\ref{fig:poli:isolation-process} shows these three components and their relationships.
\poli{} provides a unified \texttt{numpy} \citep{harris202numpy} interface, with inputs being \texttt{numpy} arrays of strings, and outputs being \texttt{numpy} arrays of floats. 
Finally, \poli{} makes sure to log every call to the black box, as well as handling evaluation budgets---\textbf{function calls are measured transparently and consistently across algorithms}.

A major issue on the endeavour of benchmarking on new problems are often conflicting package dependencies.
When evaluating a brand-new solver on an older benchmark problem, it can simply be impossible to create a Python environment in which both can run.
One may re-implement one or the other, though this is not only cumbersome but also error-prone.
With the isolation mechanisms provided by \poli{} this is no longer a problem. 
In particular, the isolation of the different components in the system (the black box, or the observer that is used for logging) allows an easy comparison to additional algorithms or on other problems, without the need to change plotting scripts.

\Cref{lst:example_experiment} shows the general workflow:\footnote{All listings in this appendix were tested using \poli{} version v1.0.1, and \polibaselines{} version v1.0.2. \href{https://colab.research.google.com/drive/1PozMS8GYIl8RB89e31b4sqtq2bqO3wO8?usp=sharing}{They can be easily implemented and tested in a Colab notebook.}} a user instantiates a black box by calling \texttt{create} with the corresponding black box's name, and the name of an observer. 
\poli{} then takes care that both black box and observer are instantiated, in different conda environments if necessary.
\texttt{create} returns an object inheriting from \texttt{Problem} which provides access to all things necessary: the black box function, initial observations, the observer and problem information like the alphabet and if sequences are aligned or not.\footnote{A future release adds a \emph{data package} attribute, attached to problems. 
The data package standardizes the information available to solvers by declaring unsupervised and supervised data; equalizing solver initialization.}
The black box is a function that takes and returns a \texttt{numpy} array, and the observer is informed automatically of any such calls. 
Each component of this workflow (black box, algorithm, and observer) can be easily exchanged by user-developed classes more specifically tailored to the need at hand. 
We refer to our online documentation\footnote{\url{https://machinelearninglifescience.github.io/poli-docs/}} for the most recent information on how to do this.

\begin{lstfloat}
\begin{python}[caption={For \poli{}-integrated problems and algorithms setting up an experiment is easy. This listing shows an example of running an experiment using the create method. In the main document, you can find another example that imports the problem factories directly.}, label=lst:example_experiment]
# the main function to instantiate problems
from poli import create  

# an example solver from poli-baselines
from poli_baselines.solvers.simple.random_mutation import RandomMutation  

# To create a problem, a user has to provide its name.
# In some cases, further parameters need to be specified.
# Optionally, the user can attach an observer through its
# name, or by using the problem.black_box.set_observer
# method.
problem = create(
    name="aloha",
    observer_name="",
    seed=0,
    observer_init_info={"CALLER": RandomMutation.__name__}
)

# The problem holds the black box, initial data,
# and other information.
f, x0 = problem.black_box, problem.x0  

# Evaluate the initial inputs if desired.
y0 = f(x0)  

# poli-baselines solvers simply take the black box ...
solver = RandomMutation(black_box=f, x0=x0, y0=y0)  

# ... and then try to solve (maximize) the problem.
solver.solve(max_iter=1000)

# If desired, an algorithm can also send information to the observer.
# problem.observer.log({"FOO", "BAR"})  
\end{python}
\end{lstfloat}

To analyze an experiment a user has to define an observer, inheriting from \texttt{AbstractObserver}.
\Cref{lst:example_observer} shows how to quickly set up a simple text-file logger.
The philosophy is to define the quantities of interest independent of problem or algorithm.
We provide standard observers, for example an \texttt{mlflow} observer, and we encourage users to implement their own, to log other metrics of interest. 
Just as black boxes, observers can run in isolation, such that the environment is independent of problem and, or solver, to facilitate consistent recording of metrics. 
Whenever observers can run directly in the same environment as the solver, they could be attached directly to the black box using the \texttt{set\_observer} method. \Cref{lst:example_observer_without_isolation} shows an example using the same workflow as above, and the \texttt{SimpleObserver} presented in \cref{lst:example_observer}.

\begin{lstfloat}
\begin{python}[caption=Observing an experiment, label=lst:example_observer]
from pathlib import Path
from uuid import uuid4
import json

import numpy as np

from poli.core.black_box_information import BlackBoxInformation
from poli.core.util.abstract_observer import AbstractObserver
from poli.core.registry import register_observer

class SimpleObserver(AbstractObserver):
    # The init and initialize_observer methods
    def initialize_observer(
        self,
        problem_setup_info: BlackBoxInformation,
        caller_info: object,
        seed: int,
    ) -> object:
        # Defining the experiment path
        self.experiment_path = caller_info["experiment_path"]

        # Saving the metadata for this experiment
        metadata = problem_setup_info.as_dict()

        # Saving the initial evaluations and seed
        metadata["seed"] = seed

        # Saving the metadata
        with open(self.experiment_path / "metadata.json", "w") as f:
            json.dump(metadata, f)

    def observe(self, x: np.ndarray, y: np.ndarray, context=None) -> None:
        # Appending these results to the results file.
        with open(self.experiment_path / "results.txt", "a") as fp:
            fp.write(f"{x.tolist()}\t{y.tolist()}\n")

if __name__ == '__main__':
    # This part needs to be done only once.
    # poli notes down the location of the current conda environment, so that if necessary, the observer can be instantiated in isolation.
    register_observer(
        observer=SimpleObserver(), observer_name="simple_observer"
    )
\end{python}
\end{lstfloat}

\begin{lstfloat}
\begin{python}[caption={Attaching an observer directly.}, label=lst:example_observer_without_isolation]
# Creating a fresh problem
problem = create(
    name="aloha",
    seed=0,
)

# Getting the black box and initial input
f, x0 = problem.black_box, problem.x0 

# Initializing an observer
observer = SimpleObserver()
observer.initialize_observer(
    problem_setup_info=f.info,
    caller_info={
      "experiment_path": Path().parent,
    },
    seed=0,
)

# Setting it
f.set_observer(observer)

# Evaluate the initial inputs if desired.
y0 = f(x0)  

# poli-baselines solvers simply take the black box ...
solver = RandomMutation(black_box=f, x0=x0, y0=y0)  

# ... and then try to solve (maximize) the problem.
solver.solve(max_iter=1000)
\end{python}
\end{lstfloat}

To expand on the suite of available black boxes, the user has to implement a subclass of the core object of \poli{}: \texttt{AbstractBlackBox}. The method to implement is \texttt{\_black\_box}, which takes as input an array of strings \texttt{x} (as well as an optional context), and outputs the result of evaluating it as an array of floats.
Listing \ref{lst:example_black_box} shows an example code snippet.
At run time, another user wanting to test an algorithm just imports the relevant black box object from \poli{}'s repository, and gets an interface to a dynamically instantiated black box potentially running in a different conda environment.
The \texttt{\_\_call\_\_} method then takes care of communication with the caller and logging to an observer. 
Some of these black boxes require additional assets (\textit{e.g.}~the weights of a neural network, or \texttt{csv} files). They are all either already provided or dynamically downloaded when the black box is used.

Optimization algorithms can be oblivious to any requirements a black box might have. Solvers have only access to the \texttt{AbstractBlackBox} interface. They communicate with the problem only indirectly via a local network socket provided by \texttt{python}'s native \texttt{multiprocessing} library.
This is the \textbf{isolation mechanism} that allows both algorithm and problem to run in different python environments.

%
%


\begin{lstfloat}
\begin{python}[caption=Implementing a black box, label=lst:example_black_box]
from string import ascii_uppercase

import numpy as np
from poli.core.abstract_black_box import AbstractBlackBox
from poli.core.black_box_information import BlackBoxInformation

class OurAlohaBlackBox(AbstractBlackBox):
    # The only method you need to define
    def _black_box(self, x, context = None):
        matches = x == np.array(["A", "L", "O", "H", "A"])
        return np.sum(matches, axis=1, keepdims=True)

    def get_black_box_info(self) -> BlackBoxInformation:
        return our_aloha_information  # Could be dynamic

our_aloha_information = BlackBoxInformation(
    name="our_aloha",
    max_sequence_length=5,
    aligned=True,
    fixed_length=True,
    deterministic=True,
    alphabet=list(ascii_uppercase),
    discrete=True,
)

f = OurAlohaBlackBox()
f(np.array([["A", "L", "O", "O", "F"]]))  # returns [[3]]
\end{python}
\end{lstfloat}

\subsection{Benchmark Introduction}
\subsubsection{Ehrlich functions}
\label{sec:appendix:ehrlich}

Ehrlich functions \citep{Stanton:Ehrlich:2024} were proposed as a closed-form alternative to black boxes like \texttt{FoldX} or \texttt{RaSP}, which could potentially raise licensing issues and, before \poli{}, were not readily available for querying. This section explains how Ehrlich functions are constructed and queried from a bird's eye view. The details can be found in the original paper.

These functions are procedurally generated from a random seed and hyperparameters like alphabet size $|\mathcal{A}|$, sequence length $L$, number of motifs $n_m$ and motif length $l$. The following are the steps that are followed to construct the problem:
\begin{enumerate}
    \item A sparse transition matrix $A$ of size $|\mathcal{A}|\times |\mathcal{A}|$ is constructed. This transition matrix spans a finite Markov chain: sequences in the problem are constructed by starting with a random symbol and following the probabilities determined by $A$. The rows of this transition matrix are probability vectors which determine how likely it is to go from one symbol in the alphabet to another. Such a transition matrix needs to satisfy that (i) the probability of going from one symbol to itself is non-zero, and (ii) it is always possible to go from any symbol to any other symbol after following a sequence of finite steps (irreducibility). In other words, we need the Markov chain spanned by $A$ to be ergodic. $A$ defines the search space of feasible sequence: the fact that $A$ is sparse means that some of the transitions are impossible.
    \item Once $A$ has been constructed, $n_m$ motifs of length $l$ are sampled according to it by sampling a sequence of length $n_m \cdot l$ and splitting it. This ensures that the motifs are feasible and easily within reach from one to the next.
    \item These motifs are to be satisfied in certain positions in the string. These positions are constructed using random offsets of size $l$. For example, the motif ``ADER'' with offsets of $[0, 1, 3, 5]$ is satisfied if the sequence contains ``ADXEXR'' where ``X'' can be any other member of the alphabet: ``D'' is at distance 1 of ``A'', ``E'' is at a distance 3 of ``A'' and ``R'' is at distance 5 of ``A''. These random offsets are randomly constructed to maximize slackness in the sequence.
\end{enumerate}

We take the original implementation\footnote{\url{https://github.com/prescient-design/holo-bench}} and wrap our black box logic around it, making it compatible with solvers in \polibaselines{}. We also include a data package made by using the original utilities for sampling the underlying transition matrix.

\subsubsection{PMO}
\label{sec:appendix:PMO}
The \textit{Practical Molecular Optimization} (PMO) benchmark contains a representative set of tasks defined on small molecule inputs with respective computational oracles. 
The input to the black box functions are alphabet representation for small molecules as either tokenized SMILES \citep{Weininger:SMILES:1988} or SELFIES \citep{Krenn:SELFIES:2020} (see below) for the exact alphabets used -- in principle one can be converted into the other.
While \textit{small} molecule sequences usually contain fewer elements in a sequence than for example proteins, the alphabet can contain more tokens making this yet another high dimensional discrete optimization space.
We build upon the work by \citet{Gao:PMOMolOpt:2022}, who propose molecular optimization focused on validity, diversity, synthesizability - using computational values for all metrics and a budget of 10.000 evaluations. The PMO suite itself is based on the existing benchmarks Guacamol \citep{Brown:Guacamol:2019}, and elements of the TDC \citep{Huang:TDC:2021}.
The types of optimization tasks for small molecules can be differentiated into: optimizing for simple metrics like QED, LogP\footnote{These are aggregate properties of small molecules, and can sometimes be poor proxies for other chemical downstream behavior.} qualitative tasks, such as scaffold hopping, rediscovery of particular substances (e.g. troglitazone), and more complex tasks such as docking surrogates and classifiers on fingerprint representations of molecules (GSK3, JNK3, DRD2) \citep{Huang:TDC:2021}.
We evaluate a selection of solvers across all tasks with multiple seeded runs on a fixed budget and report the average of the best observations, weighing all functions equally.\footnote{Equal weighting can favor methods that perform well on a few exploitable tasks (e.g. logP, QED). We draw the users attention to that fact with the alternatives to discount or discard such tasks.}
Altogether, these tasks constitute a set of functions which can be optimized and which have previously been used to assess algorithm performance in the bio-chemical domain, allowing a comparisons to previous results and reported benchmarks. 
We make these environments accessible and solvable in a unified way through the \texttt{poli} infrastructure.

\begin{lstlisting}[language=python,caption={Tokenized SELFIES alphabet used. Size=64 tokens, maximum length=70.}]
    {"[nop]": 0, "[C]": 1, "[=C]": 2, "[Ring1]": 3, "[Branch1]": 4, "[N]": 5, "[=Branch1]": 6, "[=O]": 7, "[O]": 8, "[Branch2]": 9, "[Ring2]": 10, "[=N]": 11, "[S]": 12, "[#Branch1]": 13, "[C@@H1]": 14, "[C@H1]": 15, "[=Branch2]": 16, "[F]": 17, "[#Branch2]": 18, "[Cl]": 19, "[#C]": 20, "[NH1+1]": 21, "[P]": 22, "[O-1]": 23, "[NH2+1]": 24, "[Br]": 25, "[N+1]": 26, "[#N]": 27, "[C@]": 28, "[NH3+1]": 29, "[C@@]": 30, "[=S]": 31, "[=NH1+1]": 32, "[N-1]": 33, "[=N+1]": 34, "[S@]": 35, "[S@@]": 36, "[I]": 37, "[S-1]": 38, "[=NH2+1]": 39, "[=S@@]": 40, "[=S@]": 41, "[=N-1]": 42, "[P@@]": 43, "[P@]": 44, "[NH1-1]": 45, "[=O+1]": 46, "[=P]": 47, "[=P@@]": 48, "[=OH1+1]": 49, "[=P@]": 50, "[#N+1]": 51, "[S+1]": 52, "[CH1-1]": 53, "[=SH1+1]": 54, "[P@@H1]": 55, "[=PH2]": 56, "[P+1]": 57, "[CH2-1]": 58, "[O+1]": 59, "[=S+1]": 60, "[PH1+1]": 61, "[PH1]": 62, "[S@@+1]": 63}
\end{lstlisting}

\subsection{Compute details for all experiments}
\label{sec:appendix:compute-details}

\subsubsection{Ehrlich functions}

The following solvers ran in an HPC with memory/CPU/GPU requirements given by 30G/10/1 Titan X with 12Gb of memory with: \texttt{SAASBO}, \texttt{ProbRep}, Hvarfner's \texttt{VanillaBO}, \texttt{RandomLineBO}, \texttt{DirectedEvolution}, \texttt{HillClimbing}, \texttt{CMA-ES} and \texttt{GeneticAlgorithm}. All these were time-gated via SLURM to run for at most 24hrs.

The remaining solvers (\texttt{BAxUS}, \texttt{Turbo} and \texttt{Bounce}) ran in a Google Cloud instance with one Tesla T4 with 16Gb.

\subsubsection{PMO}

\texttt{HillClimbing}, Hvarfner's \texttt{VanillaBO}, \texttt{RandomLineBO}, \texttt{SAASBO} and \texttt{ProbRep} experiments ran in an HPC cluster on CPUs using equivalent SLURM scripts (max 24h of runtime).

\texttt{Turbo} ran on an M2 Max Mac with 32Gb of memory using MPS.

\texttt{BAxUS} and \texttt{Bounce} ran on a Deep Learning compute server using the marketplace solutions of Google Cloud Platform, using a Tesla T4 (approx. 16Gb of memory).

\newpage
\section*{Checklist}


\begin{enumerate}

\item For all authors...
\begin{enumerate}
  \item Do the main claims made in the abstract and introduction accurately reflect the paper's contributions and scope?
    \answerYes{ MGD: yes, RM: yes, SB: yes, YZ: yes, SH: yes, WB: yes}
  \item Did you describe the limitations of your work?
    \answerYes{ MGD: yes, RM: yes, SB: yes, YZ: yes, SH: yes, WB: yes}
  \item Did you discuss any potential negative societal impacts of your work?
    \answerYes{ MGD: yes, RM: yes, SB: yes, YZ: yes, SH: yes, WB: yes}
  \item Have you read the ethics review guidelines and ensured that your paper conforms to them?
    \answerYes{ MGD: yes, RM: yes, SB: yes, YZ: yes, SH: yes, WB: yes}
\end{enumerate}

\item If you are including theoretical results...
\begin{enumerate}
  \item Did you state the full set of assumptions of all theoretical results?
    \answerNA{No theoretical results.}
	\item Did you include complete proofs of all theoretical results?
    \answerNA{No theoretical results.}
\end{enumerate}

\item If you ran experiments (e.g. for benchmarks)...
\begin{enumerate}
  \item Did you include the code, data, and instructions needed to reproduce the main experimental results (either in the supplemental material or as a URL)?
    \answerYes{In the project URL you can find a link to our repository with instructions.}
  \item Did you specify all the training details (e.g., data splits, hyperparameters, how they were chosen)?
    \answerYes{See Sec.~\ref{sec:appendix:training-vaes-on-selfies}.}
	\item Did you report error bars (e.g., with respect to the random seed after running experiments multiple times)?
    \answerYes{See Tables \ref{tab:results:absolute_values_for_128_latent_dim} and \ref{tab:results:absolute_values_for_2_dim_latent_space}.} 
	\item Did you include the total amount of compute and the type of resources used (e.g., type of GPUs, internal cluster, or cloud provider)?
    \answerYes{ See Sec.~\ref{sec:appendix:compute-details} in the appendix}
\end{enumerate}

\item If you are using existing assets (e.g., code, data, models) or curating/releasing new assets...
\begin{enumerate}
  \item If your work uses existing assets, did you cite the creators?
    \answerYes{We use the PMO benchmark, for which we cite not only the original developers, but also the work they are based on plus their TDC framework. Plus, we cite the authors of every HDBO solver we could find, let alone use. For Ehrlich functions, we cite the relevant paper.}
  \item Did you mention the license of the assets?
    \answerYes{ Yes, see Sec.~\ref{sec:appendix:reproducing-specific-solvers} in the appendix.}
  \item Did you include any new assets either in the supplemental material or as a URL?
    \answerYes{In the project website. \url{https://machinelearninglifescience.github.io/hdbo_benchmark/}}
  \item Did you discuss whether and how consent was obtained from people whose data you're using/curating?
    \answerNA{We are not using data on people.}
  \item Did you discuss whether the data you are using/curating contains personally identifiable information or offensive content?
    \answerNA{We are not using data on people.}
\end{enumerate}

\item If you used crowdsourcing or conducted research with human subjects...
\begin{enumerate}
  \item Did you include the full text of instructions given to participants and screenshots, if applicable?
    \answerNA{We are not crowdsourcing experiments, nor conducting research on humans.}
  \item Did you describe any potential participant risks, with links to Institutional Review Board (IRB) approvals, if applicable?
    \answerNA{We are not crowdsourcing experiments, nor conducting research on humans.}
  \item Did you include the estimated hourly wage paid to participants and the total amount spent on participant compensation?
    \answerNA{We are not crowdsourcing experiments, nor conducting research on humans.}
\end{enumerate}

\end{enumerate}


\end{document}